\documentclass[11pt]{article}

\usepackage[final]{acl}

\usepackage{times}
\usepackage{latexsym}

\usepackage[T1]{fontenc}
\usepackage[utf8]{inputenc}

\usepackage{microtype}

\usepackage{inconsolata}

\usepackage{graphicx}

\usepackage{lipsum}
\usepackage{tabularx}
\usepackage{multirow}

\title{\textsc{SinhaLegal}: A Benchmark Corpus for Information Extraction and Analysis in Sinhala Legislative Texts}

\author{Minduli Lasandi$^a$ \and Nevidu Jayatilleke$^b$ \\
  $^a$School of Computing, 
  Informatics Institute of Technology, 
  Sri Lanka \\
  $^b$Department of Computer Science \& Engineering, University of Moratuwa, Sri Lanka \\
  \texttt{minduli.20220374@iit.ac.lk, nevidu.25@cse.mrt.ac.lk}
  \small{
  }}

\begin{document}
\maketitle
\begin{abstract}
%\lipsum[80]

\textsc{SinhaLegal} introduces a Sinhala legislative text corpus containing approximately 2 million words across 1,206 legal documents. The dataset includes two types of legal documents: 1,065 Acts dated from 1981 to 2014 and 141 Bills from 2010 to 2014, which were systematically collected from official sources. The texts were extracted using OCR with \texttt{Google Document AI}, followed by extensive post-processing and manual cleaning to ensure high-quality, machine-readable content, along with dedicated metadata files for each document. A comprehensive evaluation was conducted, including corpus statistics, lexical diversity, word frequency analysis, named entity recognition, and topic modelling, demonstrating the structured and domain-specific nature of the corpus. Additionally, perplexity analysis using both large and small language models was performed to assess how effectively language models respond to domain-specific texts. The \textsc{SinhaLegal} corpus represents a vital resource designed to support NLP tasks such as summarisation, information extraction, and analysis, thereby bridging a critical gap in Sinhala legal research.

\end{abstract}

\section{Introduction}

% The limited availability of linguistic resources, combined with complex syntax structures, has made the development of tasks such as text summarisation and machine translation particularly challenging in low-resource settings. It is observed in a study by \citet{de2019survey} that only 11.43\%\ of Sinhala NLP have released the relevant datasets, 9.71\%\ of code has been released, and only 5.71\%\ of tools have been released. It was also stated that only 1.14\% have released data into public repositories. These statistics clearly indicate a significant gap in the availability of resources and the need for shareable datasets.

%Legal documents are essential for legal work, as lawyers rely on regulations, statutes, and past court decisions to address legal issues. These documents provide reliable and authoritative information \cite{pietrosanti1999advanced}. Sinhala legal texts are, however, largely inaccessible in machine-readable form because they are available only as scanned PDFs. Some of these PDFs may contain degraded print quality, making it very unclear to read. As a result, the legal domain remains one of the most resource-constrained areas within Sinhala NLP research.

Legal documentation forms the backbone of modern legal systems. These documents provide an authoritative textual basis for legislation, interpretation and judicial decision making \cite{pietrosanti1999advanced}. As a result, legal texts require a high level of precision, consistency, and well-defined structure. They commonly contain complex sentence constructions and specialised legal vocabulary that differ from those found in general-purpose texts \cite{jayatilleke2025hybrid}. These characteristics make legal documents more difficult to process automatically and highlight the need for specialised computational approaches to support various tasks.

The digitalisation of legal documents is an essential prerequisite for building reliable legal NLP systems \cite{boella2019semi}. In many such contexts, legal texts are available only in scanned or image-based formats, introducing additional challenges related to \textit{Optical Character Recognition} (OCR), layout preservation, and noise reduction. These constraints restrict broader access to legal information, reinforcing the need for systematically constructed, high-quality legal text datasets.

The Sinhala language is part of the Indo-European language family, specifically within the Indo-Aryan branch. It is the first language (L1) spoken by approximately 16 million people in Sri Lanka~\cite{de2019survey}. Sinhala features a unique script that descends from the Indian Brahmi script~\cite{fernando1949palaeographical}. Although Sinhala is classified as a large institutional language by the Ethnologue categorisation system, it is considered a low-resource language (Category 02) according to the criteria outlined by~\citet{ranathunga-de-silva-2022-languages}.

In this study, we introduce \textsc{SinhaLegal}~\footnote{\url{https://bit.ly/4buVbKx}}, a dataset that includes Sinhala legal acts and bills from 1981 CE to 2014 CE. We provide a detailed discussion of the systematic steps taken in creating this dataset, which include data collection, preprocessing, filtration, and text extraction using OCR. This process is followed by manual post-processing and concludes with the creation of metadata.

%This work aims to support future research in areas such as diachronic and synchronic analysis, legal text summarisation and other related tasks in Sinhala NLP.

\section{Related Work}

% Research in NLP relies heavily on high-quality datasets. Legal NLP, in particular, has been widely supported by numerous datasets across different languages and jurisdictions, enabling tasks such as summarisation, classification, and information extraction, and forming the foundation for much of the progress in the field.

Researchers have prioritised the development of datasets containing legislative text in various languages and jurisdictions. These datasets support tasks such as summarisation, classification, information retrieval, and both diachronic and synchronic analysis, forming the basis for significant advancements in the field.

\subsection{Sri Lanka Document Dataset}
\label{section: Sri Lanka Document Dataset}

This repository is a comprehensive collection of official Sri Lankan governmental, legal and administrative documents spanning several decades and sources from authoritative institutions \cite{senaratna2025sri}.

% The repository includes parliamentary records such as Hansard, legislation and related publications such as Acts from 1981 - 2025, Bills, and Extraordinary gazettes from 2010 - 2025. 
% A wide range of government communications, such as police press releases, Treasury announcements, and presidential Media Division releases. It also contains extensive datasets from the Disaster Management Centre. The sector-specific datasets include annual, monthly, and weekly reports from the Ministry of Fisheries on production and export–import activity. Additionally, the repository includes historic Central Bank annual reports and educational publications from the Educational Publications Department. Together, all these count up to a number of 230,091 documents across eight decades, from the 1950s to the 2020s.
The repository contains official Sri Lankan governmental, legal, and administrative documents, including parliamentary records such as Hansard, Acts (1981–2025), Bills, and Extraordinary Gazettes (2010–2025), government communications such as police press releases and Treasury announcements, documents from the Disaster Management Centre, sector-specific reports from the Ministry of Fisheries, historic Central Bank annual reports, and educational publications from the Educational Publications Department. In total, the collection comprises 230,091 documents spanning from the 1950s to the 2020s.

\textsc{SinhaLegal} focuses exclusively on the Acts and Bills contained in the Sri Lanka Document Dataset by~\citet{senaratna2025sri}. In the original repository, these legal documents are primarily available as PDF files within a much broader collection. Our study builds on this existing resource by extracting, cleaning, and structuring the Acts and Bills into a dedicated machine-readable corpus, which is further discussed in section~\ref{sec:method}.

\subsection{Cambridge Law Corpus (CLC)}

 The \textit{Cambridge Law Corpus} (\texttt{CLC}) is a substantial dataset for legal Artificial Intelligence research that comprises over 250,000 UK court cases \cite{ostling2023cambridge}. This corpus consists of 258,146 court cases drawn from 53 courts spanning over the 16th century to the 21st century %four centuries.% 
 It includes approximately 0.8 billion tokens, stored in \texttt{XML} format that captures both the full case body and rich metadata such as judge names, parties, and dates.

 During the process of creation, the word and \texttt{PDF} files were cleaned, OCR processed through the Tesseract engine \cite{kay2007tesseract} and normalised into \texttt{XML} format and iteratively refined through a cycle query-driven methodology inspired by \citet{voormann2008agile}. Due to the corpus size, only a stratified subset of 638 cases received expert-annotated outcomes.

 The \texttt{CLC} dataset has become an important benchmark for advanced legal AI tasks. It supports applications such as case outcome prediction and long-form legal text processing. Previous studies have tested models such as \texttt{RoBERTa} \cite{liu2019roberta} and \texttt{GPT-4} \cite{openai2023gpt} on this corpus, showing that long legal cases require models to handle difficult reasoning and strong semantic links across the text

\subsection{Other Legal Datasets}

Considering other legal datasets, \texttt{BIGPATENT} is one of the most influential large-scale datasets used for summarisation~\cite{sharma-etal-2019-bigpatent}. This consists of 1.3 million records of U.S. patent documents, sourced from Google Patents Public Datasets.\footnote{\url{https://bit.ly/4rSS4BN}}. Each entry pair has a full patent description and a human-written abstract (the gold-standard summary). The \textit{Japanese Tort-case Dataset} (\texttt{JTD}), the first legal judgment prediction resource for the Japanese jurisdiction, consist of 3,477 real civil judgments focused on tort cases such as defamation and privacy infringement \cite{yamada2025japanese}.

%In BIGPATENT, the summaries are highly abstractive with minimal extractive overlap, and the salient information is evenly distributed across long technical documents. This demonstrates how specialised legal-adjacent corpora can drive progress in summarisation for complex, domain-specific documents.

Extending this line of multilingual legal-NLP work, the \textit{Indian Legal Corpus} (\texttt{ILC}) by \citet{trivedi2023indian} offers 3,000+ expert-written abstractive summaries of Indian legal judgments. Similarly, \citet{nigam2025nyayaanumana} introduced \texttt{NyayaAnumana}, a large-scale Indian legal judgment-prediction dataset with 702,945 processed cases from across the judiciary. \citet{ma2021lecard} introduced \texttt{LeCaRD}, a Chinese legal case-retrieval dataset, comprising 107 query cases and over 43,000 candidate cases drawn from Supreme People’s Court criminal judgments. 

\citet{elaraby-etal-2024-adding} created a curated research subset from the \textit{Canadian Legal Information Institute} (\texttt{CanLII}\footnote{\url{https://www.canlii.org/en/}}), an open-access repository of Canadian case law, containing 1,049 long-form judicial opinions with expert-written abstractive summaries, each annotated for argument roles including Issue, Reason, and Conclusion. \citet{leitner-etal-2020-dataset} introduced a German legal Named Entity Recognition dataset under the EU (European) \textit{Lynx}\footnote{\url{http://www.lynx-project.eu/}} project, with 750 court decisions, 54,000 manually annotated entities across 19 categories. And the survey done by \citet{ariai2024natural} gives a review of the current landscape of NLP, focusing extensively on datasets and benchmarks in the legal domain.

Other datasets include \texttt{LEGAL-UQA}, the first Urdu-English legal question-answering dataset with 619 parallel question-answer pairs derived from Pakistan’s constitution \cite{faisal2024legal}; the \textit{Hindi Legal Documents Corpus} (\texttt{HLDC}) with 912,568 district court documents for bail prediction \cite{kapoor2022hldc}; the \texttt{ILDC} with 34,816 Supreme Court cases for judgment prediction \cite{malik2021ildc}; \texttt{MultiLegalPile}, a 689 GB multilingual corpus spanning 24 languages and 17 legal systems for LLM pretraining \cite{niklaus2024multilegalpile}; and \texttt{VLQA}, a Vietnamese dataset with 3,129 expert-annotated questions for legal question answering and information retrieval \cite{nguyen2025vlqa}.

% While significant progress has been made in legal NLP for high-resource languages and general NLP for Sinhala, there remains a clear gap in legal-domain resources for Sinhala. This gap motivates the present study, which focuses on developing a dataset to support Sinhala legal NLP tasks.

Collectively, existing legal NLP datasets show substantial progress for high-resource languages such as English, Chinese, German, and Hindi, supporting tasks including judgment prediction, summarisation, and question answering. In contrast, Sinhala resources are largely limited to general text collections, with no dedicated legal-domain datasets. This gap motivates the present study, which aims to develop a foundational resource for Sinhala legal NLP.

% \subsection{Other Existing Sinhala NLP Datasets}

% //Write the literature review here 

\section{Methodology}
\label{sec:method}

This methodology section describes the complete process followed to create the dataset. The workflow begins with collecting publicly available legal documents, followed by organising the files and extracting text from them using OCR. After extraction, several post-processing steps are implemented to correct errors and standardise the content. Finally, the structure of the dataset is established, and the inclusion of metadata information is discussed.

\subsection{Data Acquisition}

The initial stage involved gathering Sinhala legal documents from a publicly available repository on GitHub\footnote{\url{https://github.com/nuuuwan/lk_legal_docs}}~\cite{senaratna2025sri}, which is detailed in~\ref{section: Sri Lanka Document Dataset}. These documents were available in PDF format and contain the Sinhala version of national laws. 

At the time the repository was accessed \textit{(August 2025)}, the documents were organised into four categories: Acts, Bills, Gazettes, and Extraordinary Gazettes. The collection included 1,500+ Acts, 1,300+ Bills, 6,300+ Gazettes and 35,000+ Extraordinary Gazettes. During the data acquisition process, Gazettes and Extraordinary Gazettes were excluded because many of the PDF files had multi-column layouts and dense formatting, which are known to reduce OCR accuracy \cite{fleischhacker2025enhancing}. All the accessible Acts and Bills were downloaded to create a raw collection for further processing. In total, 2,865 PDFs were gathered. These documents covered a wide range of publication years, with Acts spanning from 1981 to 2025 and Bills from 2010 to 2025.

\subsection{Data Organisation}

All the downloaded legal documents were systematically organised to ensure consistency. For each metadata file processed, the corresponding Sinhala PDF (if available) was downloaded. Each file was saved using a descriptive and uniform naming convention automatically generated during the downloading process. The file name was constructed using three main components: the document type, publication date, and the cleaned description field \textit{(doc\_type\_date\_cleaned-description-or-id\_si.pdf)}. If the generated file name exceeded the file system length limits, it was automatically truncated during the download process to ensure compatibility.

% The extremely long file names in some documents were automatically shortened, and the documents in other languages were skipped during the process.
Documents published in languages other than Sinhala were also excluded.  If duplicate files were available, they were automatically detected when downloading and skipped to prevent redundancy. The downloaded documents were arranged into a hierarchical directory structure based on the document type (Bills or Acts) and again into subfolders based on their publication year.

\subsection{Text Extraction Using OCR}

After the document organisation process, all the 2,865 documents were processed using \texttt{Google Document AI} \footnote{\url{https://cloud.google.com/document-ai/}}  to perform OCR and extract the text from the PDF documents. A comparative study conducted by \citet{jayatilleke-de-silva-2025-zero} evaluated the performance of various OCR engines. Among the five engines tested on a synthetically created dataset for Sinhala, Surya\footnote{\url{https://github.com/VikParuchuri/surya}} emerged as the standout performer. However, when assessing a dataset of real scanned Sinhala documents, it became clear that \texttt{Document AI} achieved higher accuracy in its results \cite{jayatilleke2025sidiac}.

Since Google Document AI has a limit of 15 pages per processing request, the documents that had more than 15 pages were divided into chunks of 15 pages each. This ensured that all documents, regardless of length, were fully processed without losing any content. During this process, information such as the OCR confidence, the number of pages in the document, the number of chunks processed, document type, and published year was recorded.

After extraction, the text files were organised by publication years to maintain the chronological structure. The same filename convention used during the data acquisition was retained for consistency. This ensured that every extracted text file could easily be traced back into its original PDF and document category.

\subsection{Data Filtration}

We performed an \textit{Exploratory Data Analysis} (EDA) on all the documents to assess the dataset’s structure, distribution, and OCR quality before implementing any filtration steps. Based on the findings from the EDA, we applied several filtering steps to ensure that only high-quality, usable documents were retained for building the dataset.

\subsubsection{Exploratory Data Analysis}

% As part of the data filtration, we conducted an EDA on all the documents to understand the dataset's structure, distribution and OCR quality before applying filtration steps.

Acts and Bills were analysed separately due to differences in length and formatting. Bills were generally longer with an average of 17.2 pages, compared to 14.3 pages for Acts. OCR performance across both categories was strong, with average confidence scores of 0.967 for Acts and 0.950 for Bills.

The dataset spans 46 years, beginning in 1981. The most legislative years were identified based on the combined number of Acts and Bills. The analysis showed that
2016 had 144 documents, 2021 had 190 documents, 2022 had 175 documents, 2023 had 168 documents, and 2024 had 161 documents. This trend highlights a substantial increase in document publication in recent years.

OCR confidence values were available for all 2,865 documents. Documents were categorised into three quality levels based on their OCR confidence scores: High quality for scores above 0.8, Medium quality for scores between 0.6 and 0.8, and Low quality for scores below 0.6. Of the 2,864 documents, 2,767 (96.6\%) were classified as High Quality, 92 (3.2\%) as Medium Quality, and 5 (0.2\%) as Low.  

Based on the page count, 1,825 documents (63.7\%) were classified as Small (<10 pages), 857 (29.9\%) as Medium (11--50 pages), and only 183 (6.4\%) as Large (>50 pages). The page distributions for each category can be seen in Figure~\ref{fig: Document size}. More information on the EDA is discussed in the Appendix \ref{sec:appendix_EDA}.

\begin{figure}[h]
  \centering
  \includegraphics[width=0.98\columnwidth]{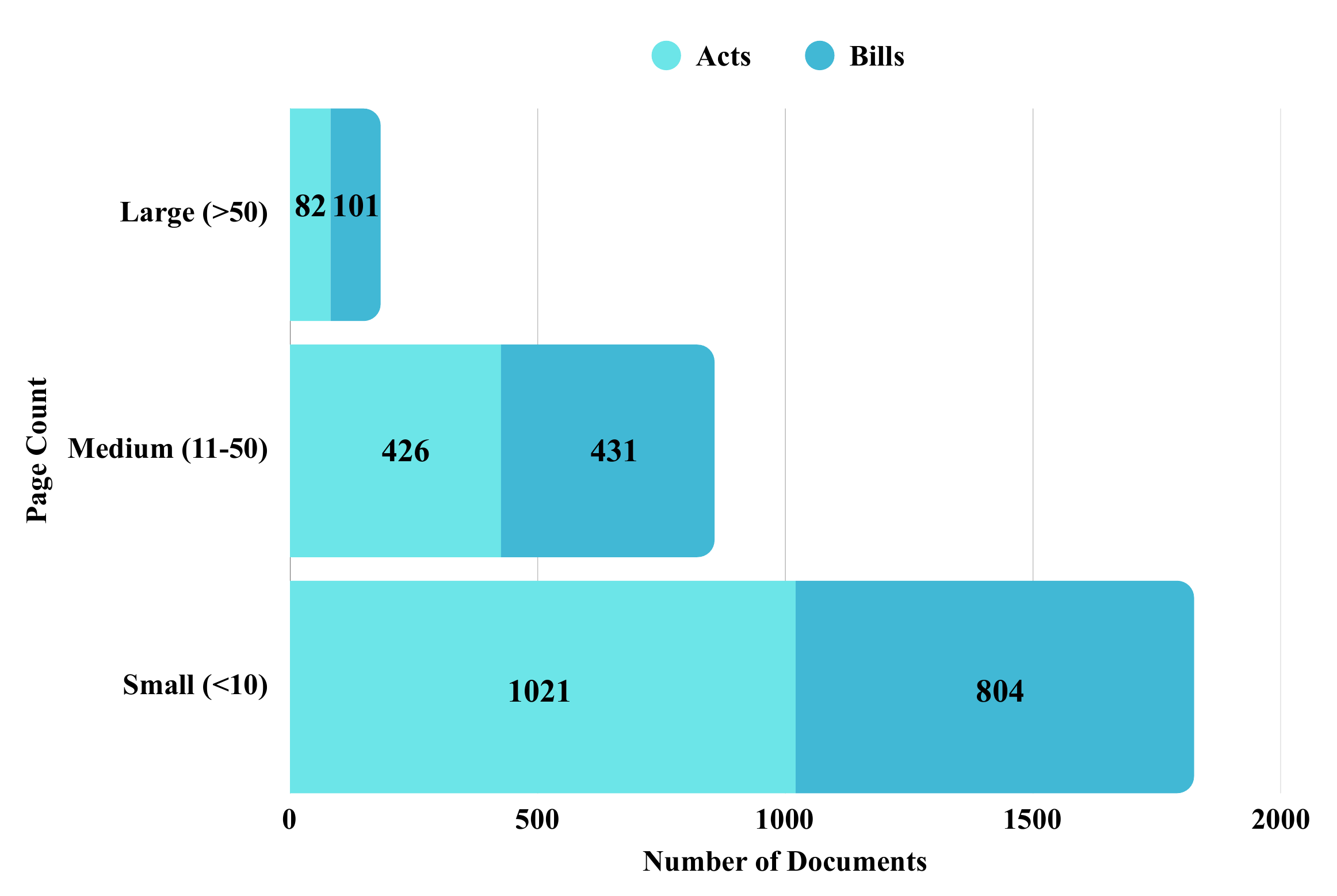}
  \caption{Distribution of document size of Acts and Bills}
  \label{fig: Document size}
\end{figure}

% The following information on the OCR quality in these documents is shown in Table~\ref{tab: OCR quality distribution}

%  \begin{figure}[h]
%   \includegraphics[width=\columnwidth]{latex/Diagrams/OCR quality by document type Radar2.PNG}
%   \caption{Distribution of OCR quality in the documents}
%   \label{fig: OCR Quality of documents}
% \end{figure}

% \begin{table}[h]
%   \centering
%   \begin{tabular}{lcc}
%     \hline
%     \textbf{OCR Quality} & \textbf{No. of documents} & \textbf{Percentage} \\
%     \hline
%     High        & 2,767 & 96.6\%\\
%     Medium     & 92 & 3.2\% \\
%     Low        & 5 & 0.2\%  \\
%           \\\hline
%   \end{tabular}
 
%   \caption{Distribution of OCR quality in the documents.}
%   \label{tab: OCR quality distribution}
% \end{table}

\subsubsection{Filtration Strategy}

% Based on the conducted EDA, several filtering steps were applied to ensure that only high-quality usable documents were retained for building the dataset. 

The downloaded documents contained a total of 2,865 legal documents, of which 1,529 were  Acts, and 1,336 were Bills. As a first step, the dataset was restricted based on publication year, retaining only Acts published between 1981 and 2014 and Bills published between 2010 and 2014, which resulted in 1,238 Acts and 155 Bills. Acts from the years 1992, 1996, and 1997, comprising 96 documents, were subsequently excluded due to visible double-sided printing that caused severe OCR errors. This ended with a count of 1,142 Acts. A page-count filter was then applied to the remaining documents, and those exceeding 50 pages were removed, as longer Acts and Bills often contained extensive tables and complex layouts that reduced OCR reliability; within the restricted year ranges, this step excluded 49 Acts and 13 Bills.

In addition to the page-count filtering, documents containing tables and multi-column layouts were removed, as these formats produced fragmented or unusable OCR text. Within the time ranges, this layout-based filtering excluded a further 26 Acts and only 1 Bill. 

\begin{figure*}[!htbp]
  \centering
  \includegraphics[width=2.0\columnwidth]{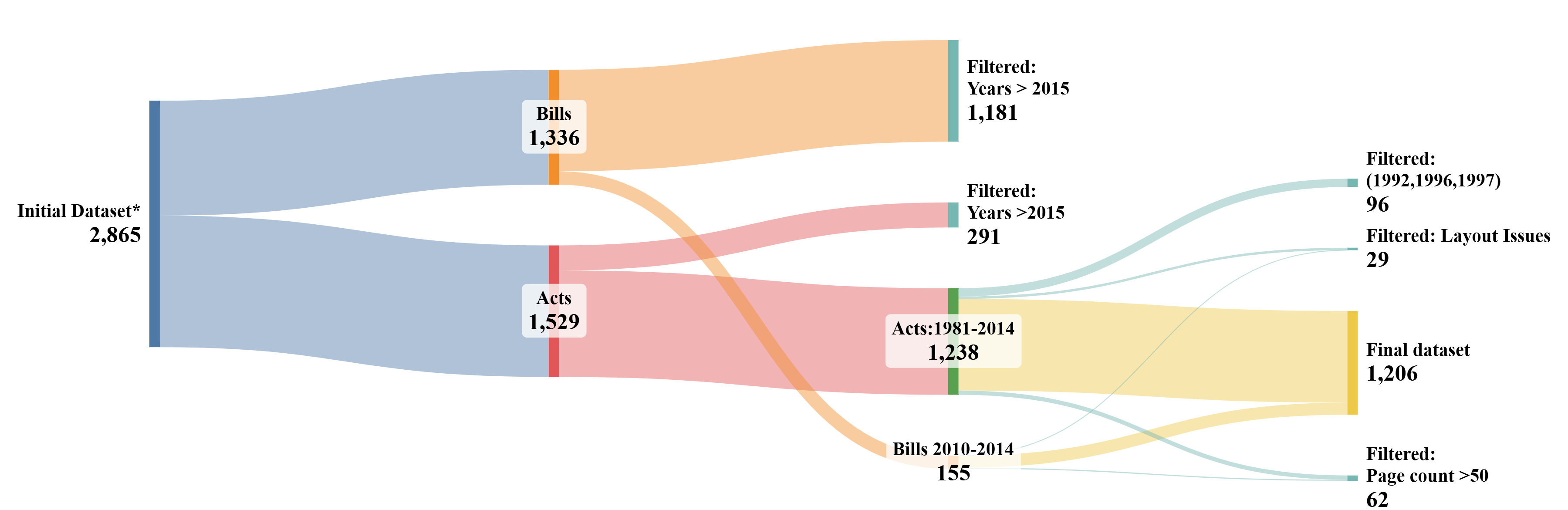}
  \caption{The flow of documents through each stage of the filtration process. *The initial dataset consists of documents that were available in the repository on the date of access (20th August 2025).}
  \label{fig: Data filtration: Sankey diagram}
  % \href{https://sankeymatic.com/build/?i=PTAEFEDsBcFMCdQDEA2B7A7gZ1AI1tBrLJKAHJoAmsWANKCgJYDWso0AFo1gFwBQIUEOHCAymgCu8AMZsA2gEEAsgHkAqmQAqAXVCaAhvADmBPnwCSkRtEb6UoACL7o%2BrAQBUcgIwBmHwDZdACFGFBQsCysbO0dnVw9QbwBWACYATl0FaWgIvhCwnG8vAA4vXSRQuHhYSh5QAB1IAE1YQxwAPlAUgAYvJL4snLl0suRKhBq6xpa20HaevrNBrG8Un2LM7N4vNNKAWgWAFjzQ8MS%2BpO18s4Xug97jvlBlnh39o7k0wIqUKsnGgAUOzSKVowP8YLSaQA7ABKMzXHC3e5eQ7eHzlcbVWoNSAABX0JlA0kkMDmSW6Ay2r12XhRh0ShwyY1%2BExxjQJRJJEjJ7QpCNOSN6dw%2Box%2BfxxABl9ABPSTQUDmLBYCQ0Kk5GnvB6JFIbcVsurSuUSBVKlVqk4FLrC%2BneQ5lCqQGKUOJuaDq7a021ebr%2BS5jJ32F0uN1mMPhgRgJqSYn6Uhu0D6chUWAAchwJPQ8HoTFY7C4vEjPCCEkoJgVAGJukloZGRPXhAA6ZtoRBJzOt0AAM07SawjEgRhQbFQmH4IBLZYI5383V0Kk4CHIsFkyusjBooArs%2D8ZkEACI5O5tPvQABbVqQHD7tRuROQGWJs%2BkhXDrsKtAANwQ%2B%2DHwEn5aJMeoCiPon4DkYuSCLebCLsSaAwPAaBnPg6AYOwaDEhIWDQGgZ6MAAXrAdbGoglC2EY8D6Ge6aJgADnRrRUZAsjNo2fD9kRoDoSU3SUqAHCgLOlJnoYRgDgwoBeCkTyINJTwKiUTy4NafC4EY8FZluXY6bpCnMVgdGGCQCpkHwkAptxUkyQJoD8qAhn6NIEGgNC%2DR4K21CIPxi4XngTwdogFbFCFoVPGgRnOdAj5eHwXZoVh8Cfs4UhsN0jbuQOHAINYXZIWeoDyv21B7AOAXIZ2FZQtVaThZF1iPulhz9CgsryoV8BeYmJp4c4jDSE8ABW2E2F2crwIw4lXuQQ0jYwY0kJQOBmaA1TfvAbiUfodGCStzguNIXAsXhdHDnAOGYZATE0O6rWoRmFVBXxz38Vw1AzQJk0cEwRgcAq6VuU8PYwF2TlsG4E1dk8TBXY5zmDqA6U2dUrU2N%2BnFsF4PpPKJRhWGNUl8Xwd2wCgTp%2BdtjGzE0TwY10xzcbAX0KtCRMkygyUoKq9FUxtoA092EhhHR1TOf2CH808dFoP2NgS6hmBPEQzOuWz%2BiodLsuMBL%2Bg9XYk2kPxWCHbAFM9UDjAbQq%2BA9tUynVPozDSwOCr9Jz3O26JCqprQjaplL1RdowAAeoCpv7DkSDpIdh%2D7vmwGgOkJkmu54IjAXp6AlDpxeLjniZ2tXnGrAyl7%2DWSww3A2GeuB2HGq780AA}{Link}
\end{figure*}

After the filtration process, 1,065 Acts and 139 Bills advanced to the next stages of research. This resulted in a total of 1,206 retained documents. A summary of the filtering stages, including the initial types of documents, page count categories, and the proportions of retained to removed documents, is presented in Figure~\ref{fig: Data filtration: Sankey diagram}. This collection of 1,206 high-quality legal documents provided a final dataset suitable for a viable post-processing procedure.

% \begin{table}[h!tb]
% \centering
% % \renewcommand{\arraystretch}{1.2}
% \resizebox{0.45\textwidth}{!}{
% \begin{tabular}{l r}
% \hline
% \textbf{Stage} & \textbf{Document Count} \\
% \hline
% \textbf{Initial Dataset} & 2,865  \\
% \quad Acts & 1,529 \\
% \quad Bills & 1,336 \\
% \hline
% \textbf{Year Range Restriction} & \\
% \quad Acts (1981--2014) retained & 1,238 \\
% \quad Bills (2010--2014) retained & 155 \\
% \hline
% \textbf{OCR Quality Filtering (Acts only)} & \\
% \quad Removed (1992, 1996, 1997) & 96 \\
% \quad Acts retained & 1,142 \\
% \hline
% \textbf{Page Count Filtering (>50 pages)} & \\
% \quad Acts removed & 49 \\
% \quad Bills removed & 13 \\
% \hline
% \textbf{Layout-based Filtering} & \\
% \quad Acts removed  & 28 \\
% \quad Bills removed & 1 \\
% \hline
% \textbf{Final Dataset Size} & \\
% \quad Acts retained & 1,065 \\
% \quad Bills retained & 141 \\
% \quad Total retained & 1,206 \\
% \hline
% \end{tabular}
% }
% \caption{Summary of data filtration stages applied to the collected legal documents.}
% \label{tab:data_filtration_summary}
% \end{table}

%Point 

\subsection{Document Post-Processing}

After OCR, several post-processing steps were performed to ensure the dataset was clean and consistent. Despite having a high accuracy score in OCR for most of the documents, the extracted text still contained structural inconsistencies that required careful cleaning. These steps were carried out manually by the authors, who are native Sinhala speakers.

The post-processing included the following corrections to address identified issues based on document-level analysis and the work by~\citet{jayatilleke2025sidiac}:

\vspace*{-1.5mm}
\paragraph{Word-level corrections:} OCR output often contained misspelt words, broken words or incorrect character substitution caused by poor quality. These were manually corrected to preserve the accuracy of the text.

\vspace*{-1.5mm}
\paragraph{Removal of footer content and page numbers:} Legal documents included footers and page numbers that disrupted the flow of paragraphs. This included the removal of such footers and page numbers.

\vspace*{-1.5mm}
\paragraph{Removal of extra sentences:} Most of the acts contained small sentences that could be seen outside the involved removing such sentences to maintain the flow and accuracy of the textragraphs. This step included removing such sentences to maintain flow, as well as the accuracy of the sentences.

\vspace*{-1.5mm}
\paragraph{Removal of seal content and prices:} This step included removing the identified watermarks that were shown as official stamps. The prices mentioned at the start of the document were also removed.

\vspace*{-1.5mm}
\paragraph{Removal of repeated titles:} Since document titles appeared multiple times per page, they were removed. The title on the title page and the first page of the document were kept.

\vspace*{-1.5mm}
\paragraph{Spacing errors:} This involved the correction of the inconsistent spaces between sentences and paragraphs.

\vspace*{-1.5mm}
\paragraph{Removal of unnecessary characters:} Occasionally, the extracted text contained characters such as underscores and dashes. These were removed since they interrupted the flow of the sentences and since they were not related to the content of the document.

\vspace{0.5em}
\noindent

These post-processing steps were conducted on all 1,065 Acts and 141 Bills in the \textsc{SinhaLegal} corpus. Appendix~\ref{sec:appendixB} provides further discussion and examples of these steps.

\subsection{Creating the Structure of \textsc{SinhaLegal}}

\begin{figure}[h]
\centering
  \includegraphics[width=0.95\columnwidth]{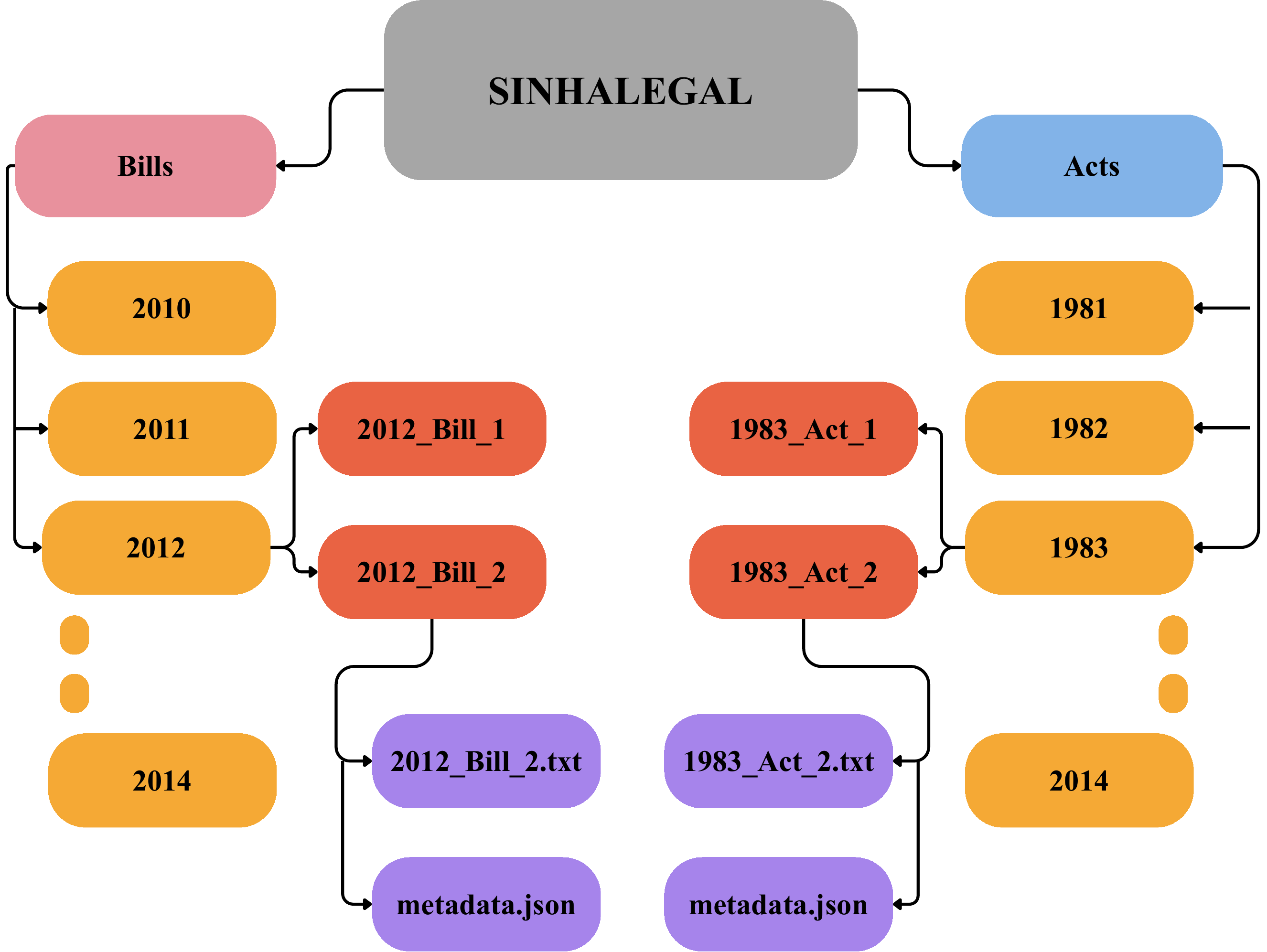}
  \caption{Structure of the \textsc{SinhaLegal} dataset}
  \label{fig: Dataset Structure}
\end{figure}

The dataset was first categorised into document type: Acts and Bills. Each document type was further organised into year-wise folders based on the year of publication. Within each year, separate directories were created for individual legal documents, with each directory named after the corresponding document. Each document directory contained the full text of the legal document and an accompanying metadata file with structured descriptive information.  An example of the composed dataset structure is depicted in Figure~\ref{fig: Dataset Structure}. Furthermore, the creation of metadata files and their records is further described in Appendix \ref{sec:appendix_metadataFiles}.

\section{Evaluation}

\subsection{Corpus Statistics}

First, we conducted a document-level evaluation of the \textsc{SinhaLegal} dataset using the entire corpus. Our analysis reveals that, on average, each document contains 1,677 word tokens, with a median length of 1,213 word tokens. The distribution of word token counts demonstrates significant variability, ranging from short texts of 95 word tokens to lengthy documents that exceed 23,000 word tokens. In total, the corpus consists of 12.8 million characters, averaging 10,678 characters per document. The figures presented in Table~\ref{fig: Dataset Structure} illustrate the heterogeneity of the legal documents.

% To provide a baseline profile of the \textsc{SinhaLegal} dataset, a document-level statistics was conducted. For this, the total dataset was considered. Based on Table \ref{fig: Dataset Structure}, it is seen that on average each document contains 1,677 words with a median length of 1,213 words. The distribution is highly variable, ranging from short texts of 95 words to lengthy documents exceeding 23,000 words. In total, the corpus comprises 12.8 million characters, with an average of 10,678 characters per document. These figures highlight the heterogeneity of legal texts.

\begin{table}[h!tb]
\centering
\resizebox{0.45\textwidth}{!}{
\begin{tabular}{lcc} % 3 columns: first left, others stretch
\hline
\textbf{Statistic} & \textbf{Value}  \\
\hline
Average word tokens per document & 1,677\\
Median word tokens per document & 1,213\\
token count range & 95 - 23,430\\
Total characters & 12,877,970\\
Average characters per document &  10,678\\
\hline
\end{tabular}
}
\caption{Summary statistics of the \textsc{SinhaLegal} dataset}
\label{tab:dataset_statistics}
\end{table}

\subsection{Lexical Diversity}
\label{subsec:lexidive}

The lexical diversity of the dataset was assessed through the \textit{Type Token Ratio} (TTR) and the distribution of hapax legomena\footnote{\label{note: def_hap}{Word types that occur only once within a given corpus.}}. Tokenisation was performed using a simple rule-based whitespace tokeniser after the non-Sinhala characters were removed using Unicode range filtering. A summary of the total word tokens, vocabulary size, and TTR for Acts, Bills, and the overall dataset is provided in Table~\ref{tab:Tokens_TTR}.

The dataset contains over two million tokens and 39,169 unique word types. TTR was length-normalised using Herdan’s C~\cite{Ross1960TypetokenM}, computed as the ratio of the logarithm of vocabulary size to the logarithm of total tokens. It is observed that the Acts account for the majority of the word tokens (1,778,265) and show a TTR of 0.7315. In contrast, Bills are smaller in size (243,942 word tokens) and demonstrate a TTR of  0.7456.

\begin{table}[h!tb]
\centering
\resizebox{0.48\textwidth}{!}{
\begin{tabular}{lccc} % 4 columns: first left, others stretch
\hline
\textbf{Statistic} & \textbf{Acts} & \textbf{Bills} &\textbf{Total Corpus}\\
\hline
Total word tokens & 1,778,265 & 243,942 &  2,022,207 \\
Vocabulary size & 37,326 & 10,390 & 39,169\\
TTR & 0.7315 & 0.7456 & 0.7284 \\

\hline
\end{tabular}
}
\caption{The number of total word tokens, vocabulary size and type token ratio for Acts, Bills and the total dataset.}
\label{tab:Tokens_TTR}
\end{table}

The analysis of hapax legomena further highlights the distribution of rare words. Across the corpus, 18,074 word types (46.14\% of the vocabulary) occur only once. Acts contain 17,632 hapax types (47.24\% of their vocabulary), while Bills contain 4,026 hapax types (38.75\%) as depicted in Table \ref{tab:hapax_legomena}. This high proportion of single‑occurrence words reflects the specialised nature of legal language, where frequent formulaic terms coexist with a long tail of rare items such as unique case names, bill titles, and technical terminology.

\begin{table}[h!tb]
\centering
\resizebox{0.3\textwidth}{!}{
\begin{tabular}{l|c|c} % 4 columns: first left, others stretch
\hline 
\multirow{2}{*}{\textbf{Documents}} & \multicolumn{2}{c}{\textbf{Hapax legomena}} \\ \cline{2-3} 
 & \textbf{Count} & \textbf{Ratio} \\ 
\hline
Acts & 17,632 & 47.24\%  \\
Bills & 4,026 & 38.75\%\\
Total corpus & 18,074 & 46.14\% \\

\hline
\end{tabular}
}
\caption{Distribution of hapax legomena across Acts, Bills, and the total corpus. }
\label{tab:hapax_legomena}
\end{table}

\subsection{Word Frequency and Coverage}
\label{sec:wordFrequency_and_Coverage}

%An analysis was conducted to identify the most repetitive words Acts and Bills. 
Word frequency analysis provides insight into the distribution of lexical items across the \textsc{SinhaLegal} corpus. Coverage statistics show that a relatively small set of high‑frequency words accounts for a substantial proportion of the text. In Acts, the top 20 words cover 23.00\% of all word tokens, while in Bills the top 20 words cover 23.32\%. Expanding to the top 50 words increases coverage to 35.04\% in Acts and 35.53\% in Bills, and the top 100 words account for nearly half of the corpus (45.89\% in Acts and 46.39\% in Bills). These figures highlight the repetitive and formulaic nature of Sinhala legal language.

It could be seen that conjunctions such as \raisebox{-0.5ex}{
\includegraphics[height=1.5\fontcharht\font`\A]{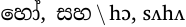}}, particle words such as \hspace*{-3pt}\raisebox{-0.5ex}{
\includegraphics[height=1.5\fontcharht\font`\A]{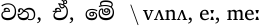}} and other terms such as \hspace*{-3pt}\raisebox{-0.5ex}{
\includegraphics[height=1.5\fontcharht\font`\A]{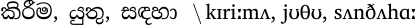}} were repeated mostly. Bills show a similar distribution, with high coverage by a similar set of words. The top 10 identified frequent words are discussed in Appendix \ref{sec:appendix_WordFrequency}.

The coverage statistics demonstrate that Sinhala legal texts rely heavily on a small core vocabulary, while still maintaining lexical breadth through lower‑frequency items. This shows that these documents are highly standardised in their functional framing, yet expansive in their incorporation of specialised terminology.

\subsection{Calculating Text Accuracy and Structure}

This dataset was evaluated using character-level and word-level error metrics, following a similar approach to CLC~\cite{ostling2023cambridge}. For this, the corrected text was taken as the ground truth. Word Error Rate (WER) and Character Error Rate (CER) were computed with and without text normalisation. The results show WERs of 26.87\% and 23.44\%, and CERs of 24.07\% and 24.06\%, respectively. Structural differences such as line breaks were also analysed; further details are provided in Appendix \ref{sec:appendix_WER_CER} and \ref{sec:appendix_structuralAnalysis}.

\subsection{Named Entity Recognition}

A rule-based \textit{Named Entity Recognition} (NER) was implemented to identify salient entities in this dataset. Although various libraries exist for NER, they are not domain-specific and are incompatible for the legal domain \cite{badji2018legal}, especially considering Sinhala. Therefore, a rule-based approach was implemented in \texttt{Python}, utilising regular expression matching and keyword-based rules to identify legal entities. This approach was designed to capture six major types of entities:

\vspace*{-1.5mm}
\paragraph{Date:}Since dates are central to legal documents, the years and date expressions were captured using digit-based patterns (e.g., \texttt{\textbackslash b\textbackslash d {4\}\textbackslash b}}) and extended rules for textual date formats.

\vspace*{-1.5mm}
\paragraph{Title:} Institutional roles and titles such as \hspace*{-3pt}\raisebox{-0.5ex}{ \includegraphics[height=1.5\fontcharht\font`\A]{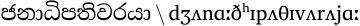}}\hspace*{-2pt} (President) and \hspace*{-3pt}\raisebox{-0.5ex}{ \includegraphics[height=1.5\fontcharht\font`\A]{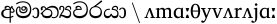}} (Minister) were identified and listed. This ensured capturing references to officials and positions within the legal system.

\vspace*{-1.5mm}
\paragraph{Organisation:} These were extracted using a keyword dictionary of institutional terms frequently occurring in Sinhala legal texts such as \hspace*{-3pt}\raisebox{-0.5ex}{ \includegraphics[height=1.5\fontcharht\font`\A]{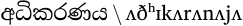}} (court of justice), \hspace*{-3pt}\raisebox{-0.5ex}{ \includegraphics[height=1.5\fontcharht\font`\A]{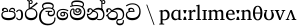}} (parliament).

\vspace*{-1.5mm}
\paragraph{Law:} Law names are highly formulaic, often ending with the word \raisebox{-0.5ex}{ \includegraphics[height=1.3\fontcharht\font`\A]{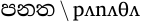}} (`act'), and to avoid false positives, common prefixes such as \hspace*{-3pt}\raisebox{-0.5ex}{ \includegraphics[height=1.5\fontcharht\font`\A]{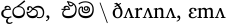}} (bear with, that same), were removed.

\vspace*{-1.5mm}
\paragraph{Person:} Personal names in legal texts are typically followed by honorifics such as \hspace*{-3pt}\raisebox{-0.5ex}{ \includegraphics[height=1.5\fontcharht\font`\A]{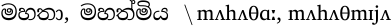}}\hspace*{-7pt} (Mister, Miss). This was used to capture up to two preceding words, ensuring that both single and compound names were recognised.

\vspace*{-1.5mm}
\paragraph{Amount:}Monetary values are expressed with numerals followed by currency markers, such as \hspace*{-3pt}\raisebox{-0.5ex}{ \includegraphics[height=1.5\fontcharht\font`\A]{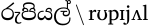}} or the letter \raisebox{-0.5ex}{ \includegraphics[height=1.5\fontcharht\font`\A]{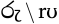}} (rupees),  ensuring accurate identification of financial references. 

\vspace{0.5em}
\noindent
The pipeline extracted a total of 28,937 entities across the corpus, and the frequencies are depicted in Table \ref{tab:NER_Results}. It could be seen that, on average, each document contains 24 entities, with a maximum of 361 and a minimum of 4 for a document, which highlights the density of entities in the corpus.

\begin{table}[h!tb]
\centering
\resizebox{0.40\textwidth}{!}{
\begin{tabular}{lc} 
\hline
\textbf{Entity type} & \textbf{Number of Entities}  \\
\hline
Date [DATE] & 13,532 \\
Title [TITLE]  & 8,736 \\
Organisations [ORG] &  4,281 \\
Law [LAW]  & 2,255\\
Person [PERSON] &  126\\
Amount [AMOUNT]& 7 \\
\hline
\end{tabular}
}
\caption{Number of entities extracted from the \textsc{SinhaLegal} dataset}
\label{tab:NER_Results}
\end{table}

\subsection{Topic Modelling}

We performed topic modelling to explore the thematic structures within the dataset using \textit{Latent Dirichlet Allocation} (LDA) \cite{blei2003latent}, and it was implemented using the \texttt{Gensim} library in \texttt{Python}. Prior to this implementation, we performed standard preprocessing steps, including tokenisation and removal of stop words.

The corpus was preprocessed to ensure consistency and reduce noise. Texts were tokenised into word units, normalised to reduce orthographic variation \cite{manning2008introduction}. The Sinhala stop words were taken from an available public GitHub~\footnote{\url{https://bit.ly/4tihUQj}} repository that was created by \citet{lakmal-etal-2020-word}, and later modified manually with the common stop words in the \textsc{SinhaLegal} dataset.  

Topic coherence was computed to analyse the model's behaviour across different values of K, with k=15 achieving the highest value. However, prior studies have shown that coherence alone is insufficient for determining the optimal number of topics, as larger values of K may lead to over-clustering and unstable topic solutions~\cite{greene2014many}. Therefore, the topic model was trained with ten topics to balance interpretability and coverage \cite{griffiths2004finding}.

The results revealed recurring themes centred on the legislative acts (\hspace*{-3pt}\raisebox{-0.5ex}{ \includegraphics[height=1.3\fontcharht\font`\A]{Sinhala_Words/si_word_panatha.pdf}}), institutional references such as courts (\hspace*{-3pt}\raisebox{-0.5ex}{ \includegraphics[height=1.5\fontcharht\font`\A]{Sinhala_Words/si_word_adhikaranaya.pdf}}), themes related to money (\hspace*{-3pt}\raisebox{-0.5ex}{ \includegraphics[height=1.5\fontcharht\font`\A]{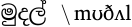}}), pension (\hspace*{-3pt}\raisebox{-0.5ex}{ \includegraphics[height=1.5\fontcharht\font`\A]{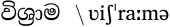}}), commissions (\hspace*{-3pt}\raisebox{-0.5ex}{ \includegraphics[height=1.5\fontcharht\font`\A]{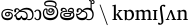}}) and elections (\hspace*{-3pt}\raisebox{-0.5ex}{ \includegraphics[height=1.5\fontcharht\font`\A]{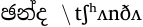}}). Representative word distribution for each topic is provided in the Appendix \ref{sec:appendix_TopicModelling}.

\subsection{Evaluation of Language Models}
\label{sec: calculating_Perplexity_Scores}

Perplexity is a standard evaluation metric in language modelling that measures how well a model predicts the next token \cite{meister-cotterell-2021-language}. Perplexity was used to evaluate how well different language models handle legal domain text, and the scores were compared with a general Sinhala dataset named \texttt{MADLAD CulturaX}~\footnote{\url{https://huggingface.co/datasets/polyglots/MADLAD_CulturaX_cleaned}} \cite{aravinda2025sinllama}.

To calculate the perplexity, we created balanced evaluation samples. The \textsc{SinhaLegal} dataset was divided into sentences and clustered using the \texttt{KMeans} algorithm with k set to 10. We selected 200 sentences from each of the 10 clusters, resulting in a total of 2,000 sentences for our analysis. This clustering step ensured that sentences were grouped by similarity, allowing us to sample proportionally from each cluster. This approach preserved diversity in terms of sentence length, topics, and overall coverage. 

For the general Sinhala dataset, which contained 10 million sentences, we randomly selected a sample of 100,000 sentences. We then applied the same clustering method as used for the legal dataset, ultimately extracting another set of 2,000 sentences.

To compare the differences between general Sinhala and legal Sinhala corpora, we selected two subsets of 2000 sentences each, created using \texttt{MADLAD CulturaX} and \textsc{SinhaLegal}. Each sentence was tokenised using a custom Sinhala tokeniser that removes non-Sinhala characters and splits text into tokens, as detailed in subsection~\ref{subsec:lexidive}. We then computed normalised word distributions for both corpora and measured their divergence using the \textit{Jensen-Shannon divergence} (JSD) \cite{lin2002divergence}, which quantifies the similarity between two probability distributions.

The computed JSD between the legal and general Sinhala corpora was 0.614, indicating a substantial difference in their word distributions. This confirms that legal Sinhala employs a distinct vocabulary and word usage compared to general language, complementing the perplexity-based evaluation and providing a quantitative measure of domain-specific linguistic characteristics.

% For the model selection, 3 large language models and 2 small language models were selected. The LLMs chosen were \texttt{Deepseek-1.3B}~\footnote{\url{https://huggingface.co/deepseek-ai/deepseek-coder-1.3b-instruct}}, \texttt{Mistral-7B}~\footnote{\url{https://huggingface.co/mistralai/Mistral-7B-v0.1}} \cite{jiang2023mistral7b}, and \texttt{Falcon-7B}~\footnote{\url{https://huggingface.co/tiiuae/falcon-7b}} \cite{almazrouei2023falcon}, all of which represent modern transformer architectures. As SLMs, \texttt{DistilGPT‑2}~\footnote{\url{https://huggingface.co/distilbert/distilgpt2}}, a distilled variant of \texttt{GPT‑2} designed for lightweight deployment, and \texttt{Gemma‑2B}~\footnote{\url{https://huggingface.co/google/gemma-2b}}  \cite{team2024gemma}, a compact model released by Google that offers competitive performance in resource‑constrained environments, were selected. This mix of models examined how size and architecture influence perplexity across legal and general Sinhala corpora.

For the perplexity-based evaluation, we considered several modern transformer architectures that support Sinhala, including \texttt{Llama-3.1-8B}~\footnote{\url{https://bit.ly/49Ieoa4}}\cite{kassianik2025llama}, \texttt{Mistral-7B}~\footnote{\url{https://bit.ly/49nZLHD}} \cite{jiang2023mistral7b}, and \texttt{Falcon-7B}~\footnote{\url{https://huggingface.co/tiiuae/falcon-7b}} \cite{almazrouei2023falcon}, \texttt{Deepseek-1.3B}~\footnote{\url{https://bit.ly/3N76iiC}}\cite{guo2024deepseek}, \texttt{DistilGPT‑2}~\footnote{\url{https://huggingface.co/distilbert/distilgpt2}}, a distilled variant of \texttt{GPT‑2}. We also included \texttt{Gemma‑2B}~\footnote{\url{https://huggingface.co/google/gemma-2b}}  \cite{team2024gemma} a compact model released by Google that demonstrates competitive performance in resource-constrained environments. This diverse selection of models allowed us to examine how the size and architecture influence perplexity across legal and general Sinhala corpora.

\begin{table}[h!tb]
\centering
\resizebox{0.45\textwidth}{!}{
\begin{tabular}{lcc} % 3 columns: first left, others stretch
\hline
\textbf{Model Name} & \textbf{\texttt{MADLAD CulturaX}} & \textsc{\textbf{SinhaLegal}}  \\
\hline
\texttt{Llama-3.1} & \underline{3.05} & \textbf{2.55}\\
\texttt{Deepseek-1.3B} & 3.30 & 2.94\\
\texttt{Mistral‑7B} & 3.68 & 3.18\\
\texttt{Falcon‑7B} &  \textbf{2.77} & \underline{2.61}\\
\texttt{DistilGPT‑2} & 6.45 &  5.77\\
\texttt{Gemma‑2B} & 8.75 & 5.59\\

\hline
\end{tabular}
}
\caption{Comparison of perplexity scores of the two datasets. \textbf{Bold:} indicates best performance and \underline{Underline:} indicates the second best.}
\label{tab:perplexityScores}
\end{table}

During the evaluation, all models exhibited lower perplexity scores on the \textsc{SinhaLegal} corpus in comparison to the \texttt{MADLAD CulturaX} dataset. This suggests that domain‑specific legal text is more predictable than general cultural content. Even though legal terms are more complex than general Sinhala, lower perplexity can likely occur due to repetitive structures and frequent patterns in texts \cite{yao-etal-2025-understanding}. Frequent phrases such as "\hspace*{-3pt}\raisebox{-0.5ex}{ \includegraphics[height=1.5\fontcharht\font`\A]{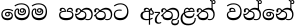}}" meaning "this act contains" (\hspace*{-7pt}\raisebox{-0.5ex}{ \includegraphics[height=1.4\fontcharht\font`\A]{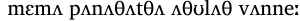}}), can be seen multiple times across documents. This repeated usage, along with the findings on the non-uniformity of word frequencies discussed in subsection~\ref{sec:wordFrequency_and_Coverage}, likely contributes to the lower perplexity scores observed when compared to the general Sinhala dataset. %The most frequently used words, as mentioned in Section \ref{sec:wordFrequency_and_Coverage}, can be seen in  Appendix \ref{sec:appendix_WordFrequency}. 

% The results revealed clear distinctions between the models. 
\texttt{Llama 3.1} and \texttt{Falcon‑7B} achieved the lowest perplexity on both datasets, followed by \texttt{Deepseek-1.3B}, indicating strong predictive performance. \texttt{Mistral-7B} also performed competitively, with slightly higher perplexity scores. As expected, the smaller models exhibited high perplexity values, in which \texttt{DistilGPT‑2} produced moderate scores, while \texttt{Gemma‑2B} showed the highest perplexity, particularly on the general corpus as presented in Table~\ref{tab:perplexityScores}.

\section{Conclusion}

This study introduced \textsc{SinhaLegal}, a Sinhala legal dataset designed to support research in legal NLP and information extraction tasks, specifically facilitating diachronic analysis of legal documents. The dataset includes a total of 1,206 legal documents, of which 141 are Bills ranging from 2010-2014 CE, and 1,065 are Acts ranging from 1981-2014 CE. The process of creating this dataset included performing OCR, filtering unwanted documents, and post-processing them manually to reduce noise and improve quality. The conducted evaluation included the lexical diversity, word frequency and coverage, NER and topic modelling to identify the number of entities and topics within the dataset. Finally, the perplexity scores were measured on selected language models to see how well the models respond to domain-specific data.   

%For future work, this dataset can be enhanced with more legal documents other than Acts and Bills. Improving the context by extending more processing methods, such as sectioning the documents, can also be done. 

For future work, this dataset can be expanded with additional types of legal documents beyond Acts and Bills. Its utility can also be enhanced by applying further post-processing methods, such as segmenting documents into sections. \textsc{SinhaLegal} fills a significant gap in legal NLP for Sinhala and provides a reliable foundation for future research.

\section*{Limitations}

\paragraph{Scope restricted to Acts and Bills:} This study considers only Acts and Bills. But the new repository in GitHub~\footnote{\url{https://github.com/nuuuwan/lk_datasets}}  mentioned in section \ref{section: Sri Lanka Document Dataset} has been updated later and contains many additional categories.

\paragraph{Temporal coverage for Acts and Bills:} Although Acts in the repository span a broader range from 1981 to 2025, and Bills a range from 2010-2025, this analysis consists of Acts and Bills published between 2014.

\paragraph{Document structure not explicitly segmented:} While some documents contain section boundaries, the documents in the dataset are provided as continuous text and are not consistently segmented into structural sections (e.g., preamble, definitions, clauses, schedules).

\paragraph{Language coverage limited to Sinhala:} Although official English and Tamil versions of legal documents, including Acts and Bills, were available, this study focuses exclusively on the Sinhala versions of the documents.

\paragraph{Manual evaluation of the NER task:} Due to the language-specific characteristics of named entities, automated evaluation methods were not fully applicable. Consequently, the NER task can be considered for manual evaluation.

\paragraph{Consideration of lengthy documents:} For practical reasons for manual post-processing, documents longer than 50 pages were not considered in this study. They can be considered for future expansion of this dataset.

\bibliography{custom}

\appendix

\section{Exploratory Data Analysis}
\label{sec:appendix_EDA}

\begin{figure*}[!htbp]
  \centering
  \includegraphics[width=2.0\columnwidth]{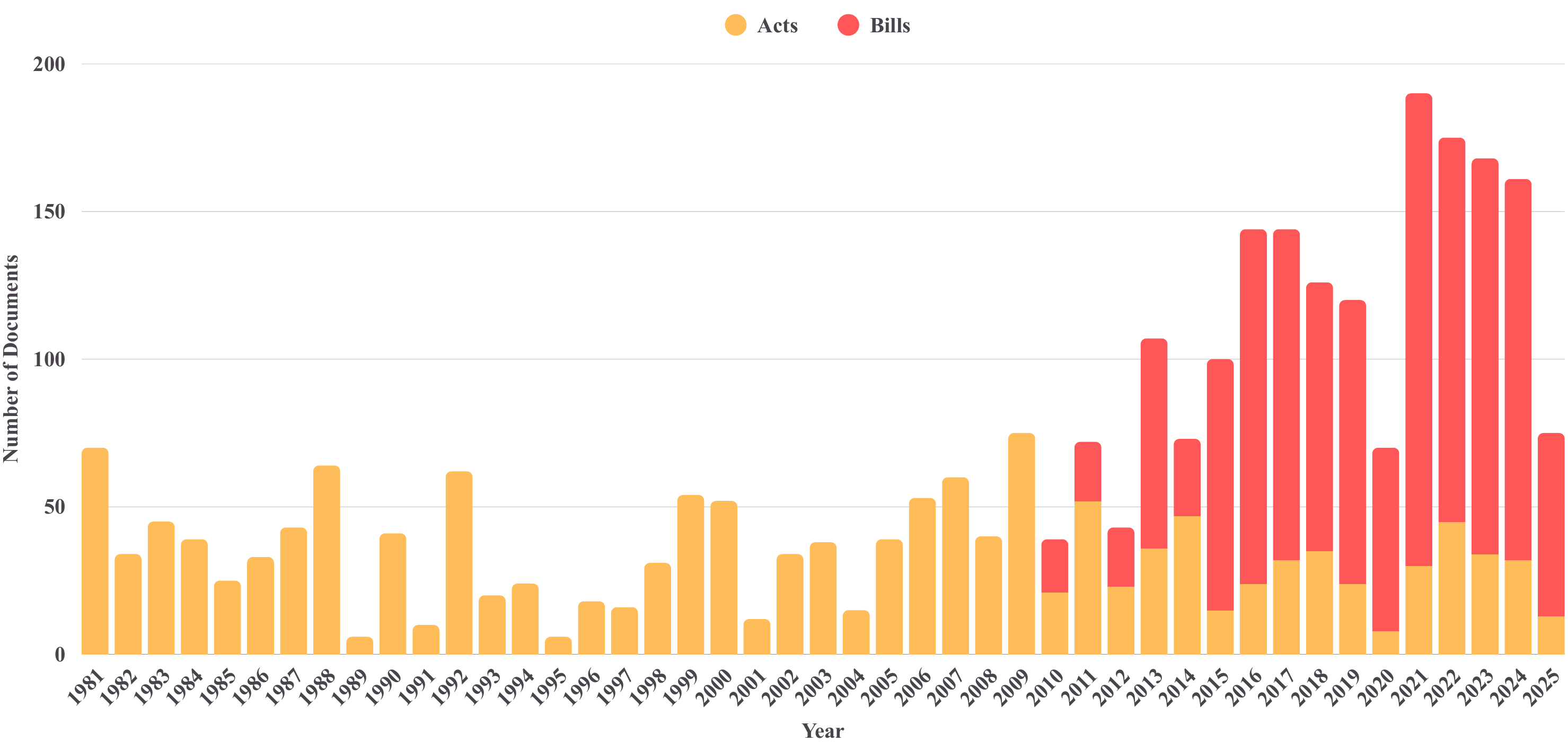}
  \caption{Distribution of Acts and Bills throughout the years from 1981-2025}
  \label{fig: Yearly distribution_Documents}
\end{figure*}

 Figure \ref{fig: Yearly distribution_Documents} presents the yearly distribution of the documents, separated by Acts and Bills. This shows a clear increase in legislative document production in recent years, particularly after 2010. While Acts are consistently present throughout the entire time span, Bills become more prominent in later years. 

The stacked representation further shows that Bills contribute significantly to the overall document count in peak years such as 2021, 2022 and 2023. Overall, this figure highlights how legal documentation has evolved, with an increasing volume and complexity in recent years, potentially reflecting changes in governance, policy focus, or administrative practices.

\begin{figure}[h]
  \includegraphics[width=\columnwidth]{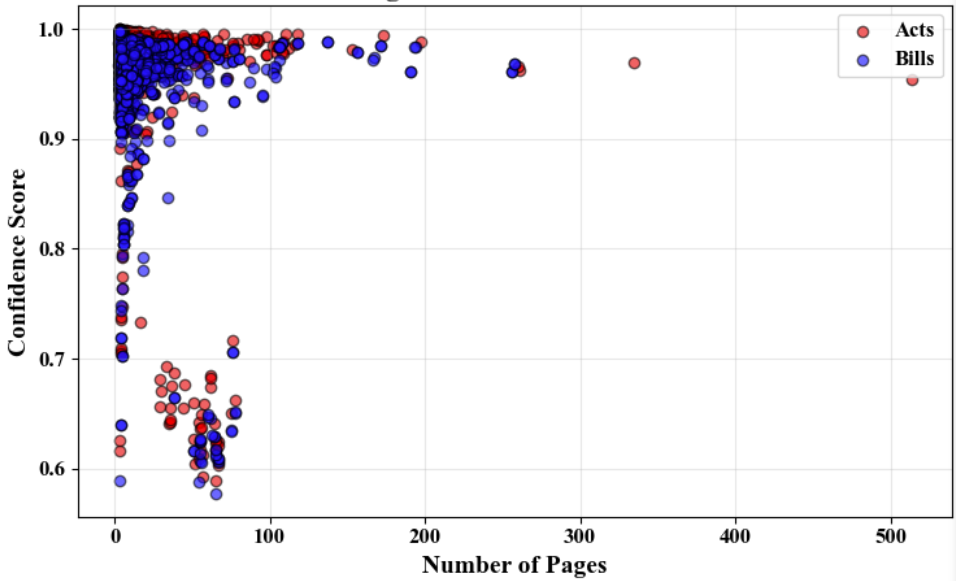}
  \caption{Relationship between the page count and OCR confidence}
  \label{fig:ocr_pagecount}
\end{figure}

The scatter plot shown in Figure \ref{fig:ocr_pagecount} depicts the relationship between document length (number of pages) and the OCR confidence scores for both Acts and Bills. It was seen that the shorter documents generally exhibited high OCR confidence values, often exceeding 0.9. As document length increases, greater variability in OCR confidence can be observed, mostly for documents exceeding 50 pages.

Longer documents frequently contain complex layouts such as multi-column formatting and tables. This can negatively impact the performance of OCR and lead to noisy text extraction. This pattern directly informed the page-count-based filtration criteria applied during dataset construction.

\begin{figure}[h]
  \includegraphics[width=\columnwidth]{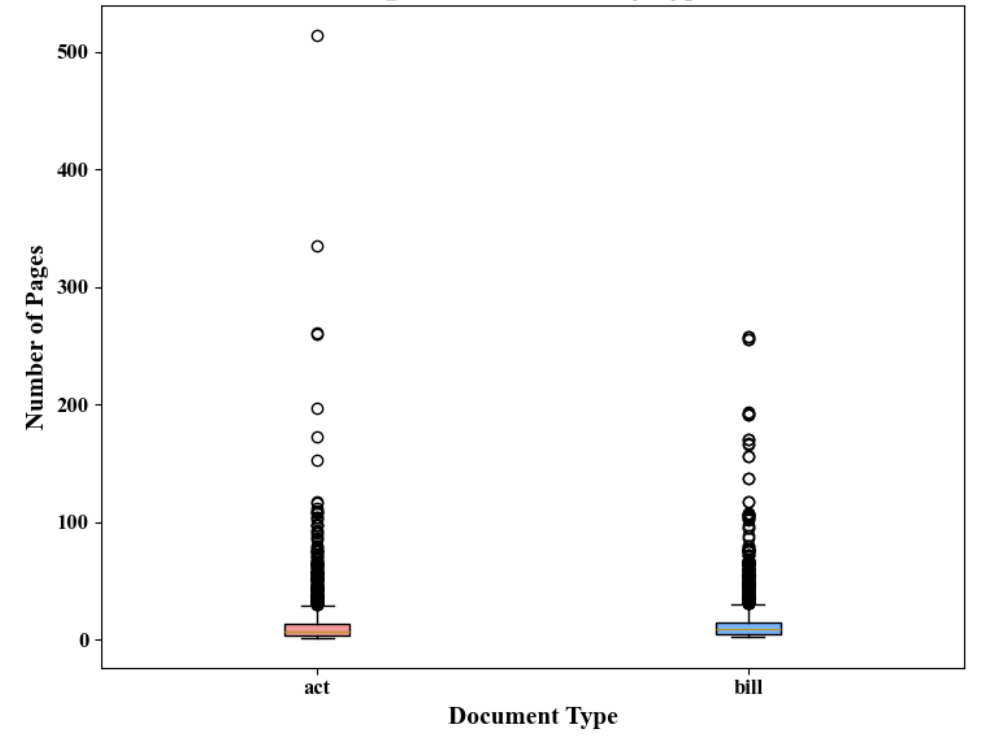}
  \caption{Boxplot showing the distribution of page counts for legal documents by type.}
  \label{fig:boxPlot}
\end{figure}

Figure \ref{fig:boxPlot} illustrates the distribution of page counts for Acts and Bills. Both document types generally have a low median page count, indicating that most acts and bills are relatively short. However, there is a notable presence of outliers, particularly among acts, with some extending beyond 500 pages. This suggests that while the majority of these documents are concise, acts tend to vary more widely in length and can be significantly longer than bills. The variability in page count highlights the diverse complexity and scope of legislative documents within these categories.

\section{Document Post-Processing}
\label{sec:appendixB}

Seven errors were identified and addressed during the post-processing phase. This was carried out by the author, who is a native Sinhala speaker and is fluent in Sinhala

The acts contained small sections of sentences that could be seen next to paragraphs. These were often seen to be breaking the flow of the paragraphs and creating low accuracy of the meaning of the mentioned text. Hence, these were removed. Some examples are shown in Table~\ref{tab:extra_sentences}.

%Removal of extra sentences: table-%
\begin{table}[h!tb]
\centering
\resizebox{0.45\textwidth}{!}{
\begin{tabular}{lc}
\hline
\textbf{Extra Sentences} & \textbf{Extracted Text} \\
\hline
\includegraphics[width=0.45\columnwidth]{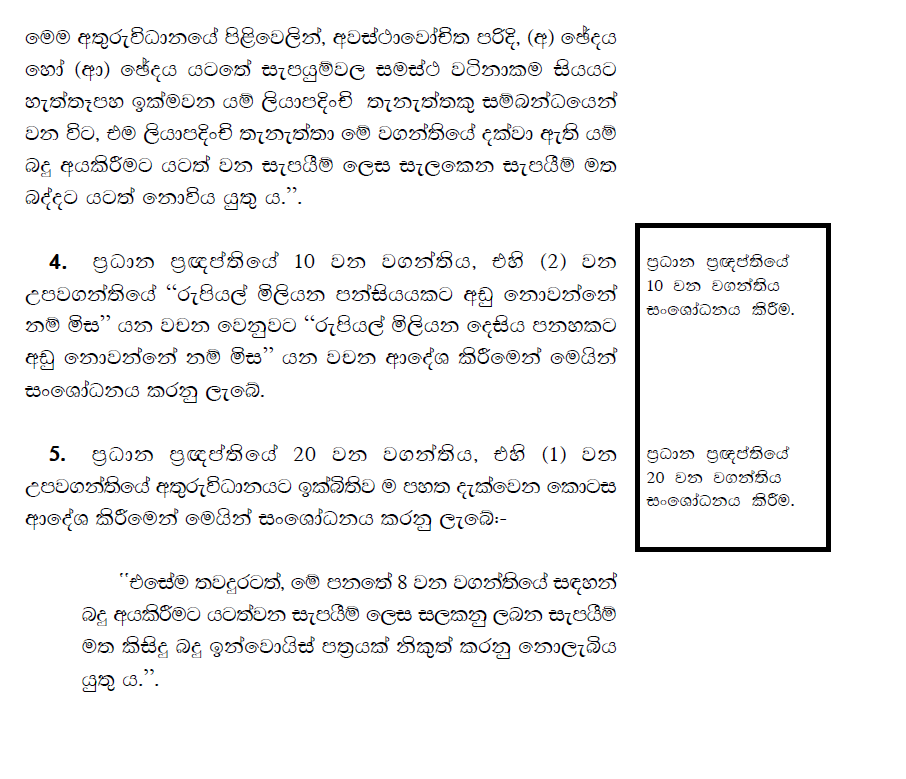} &
\includegraphics[width=0.45\columnwidth]{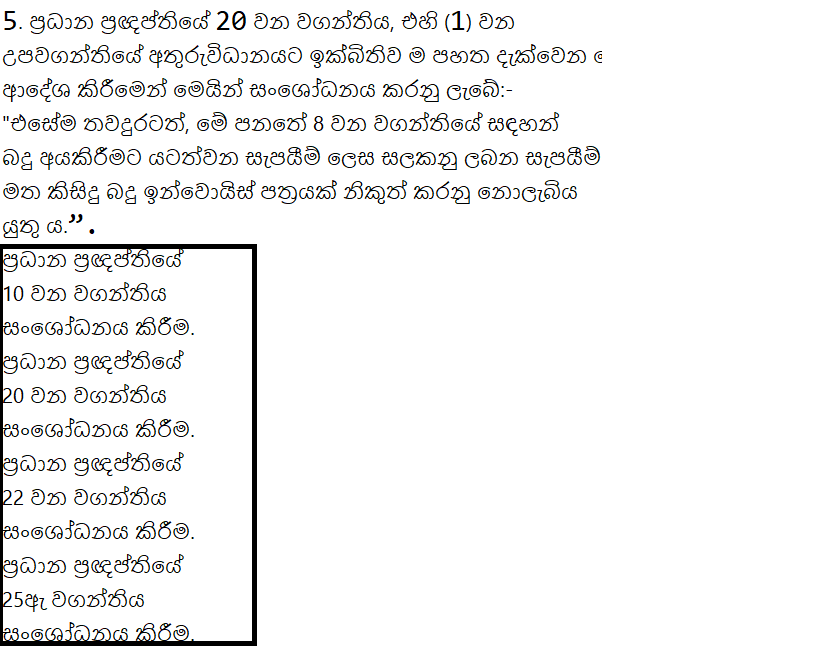} \\ \hline

\includegraphics[width=0.45\columnwidth]{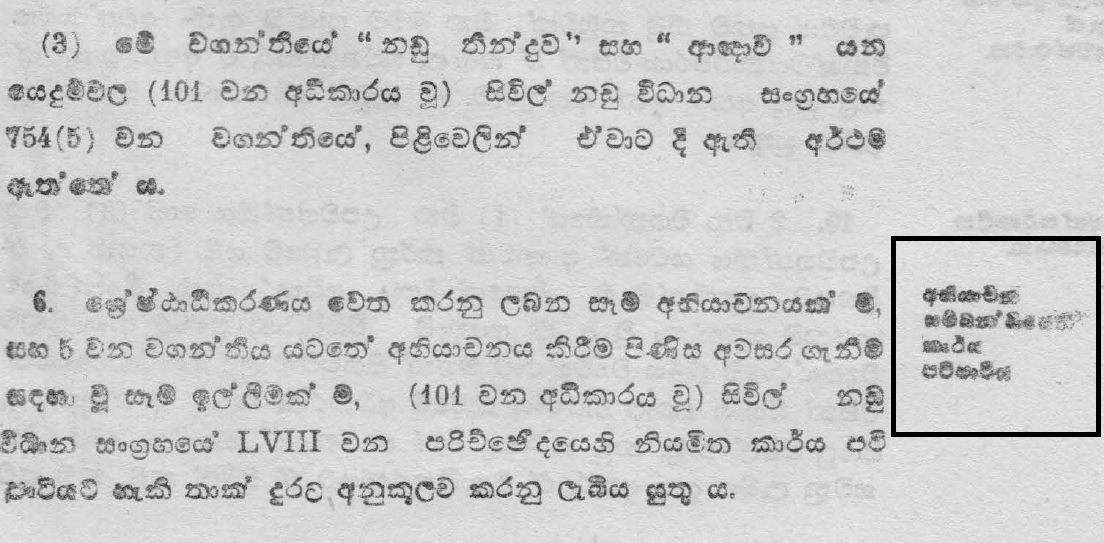} &
\includegraphics[width=0.45\columnwidth]{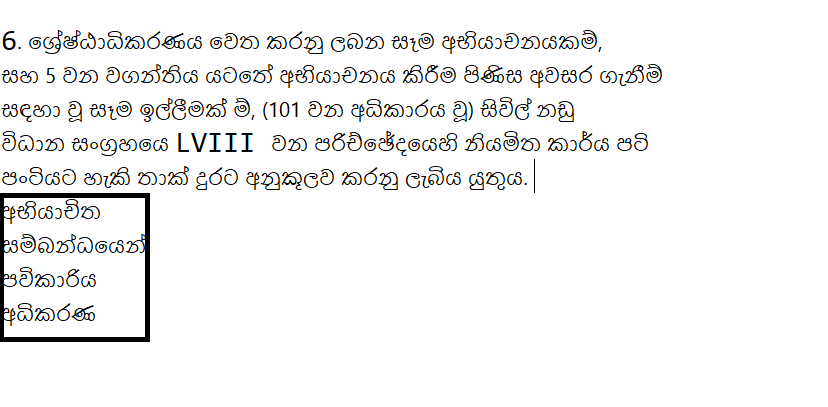} \\ \hline

\end{tabular}
}
\caption{Examples of extra sentences in the scanned PDF and extracted text removed during post-processing.}
\label{tab:extra_sentences}
\end{table}

Some of the years contained low-quality scanned PDFs, hence the output contained fragmented and misspelt words. Several paragraphs were seen to be broken, and this required manually correcting the spelling and order of the sentences. Some words could not be identified at all, and they too had to be manually typed and added into the extracted text. This took the most amount of time compared to the other errors that were fixed. Some of the above-mentioned errors and the corrected version can be seen in Table~\ref{tab:word_level_corrections}. 

These documents also contained footers and page numbers that were not relevant to the legal content of the document. The page numbers were normally present at the end of each page, and the footers at the end of each document or section. Some of the Acts could be seen with seal content, which also does not have an impact on the content present in the legal document. Hence, these seal contents were removed from the documents. The titles of the document could be seen repeatedly on every page of the document. The title on the title page (the very first page) and the first page of the document were kept, and the others were removed to keep the flow of the document without the titles interrupting the paragraphs. 

A large portion of the document exhibited inconsistent spacing. In some cases, excessive blank spaces appeared between lines, while in others, paragraphs, numbered lists and bullet points were merged with no spacing at all. These inconsistencies made the text visually congested and difficult to read. During post-processing, proper line breaks and spacing were restored between paragraphs, numbered lists and bullet points to ensure a clear and well-structured document layout. An example of the mentioned spacing error and its respective corrections can be seen in Table~\ref{tab:spacing_corrections}.

The extracted text also contained various unnecessary characters produced by OCR misidentification, including underscores, dashes, semicolons, brackets, random English letters and other stray symbols \textit{(e.g.,\texttt{\_; - \{ :] /)}}. These artefacts appeared randomly throughout the text and did not carry any semantic meaning. All such characters were removed during post-processing to ensure clean and consistent documents.

\section{Creating Metadata Files}
\label{sec:appendix_metadataFiles}

During the OCR process, document-level information was extracted and recorded for each legal document. Following post-processing, the relevant information related to each document was taken into separate metadata files and grouped accordingly. Maintaining document-level metadata also supports reproducibility and auditability, allowing OCR results and evaluation metrics to be traced back to their original source documents.

\begin{figure}[h]
  \includegraphics[width=\columnwidth]{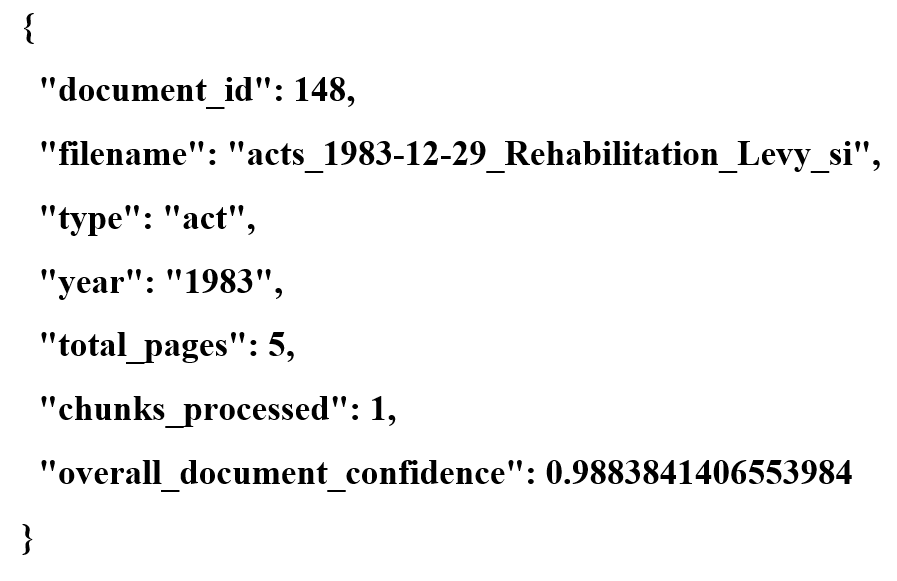}
  \caption{Example of the metadata record for a document}
  \label{fig: Metadata record structure}
\end{figure}

The metadata files consisted of the document ID, the file name, the document type (Acts/Bills), the year of publication, the total number of pages, the number of chunks (1 chunk = 15 pages maximum) processed during the OCR process and the overall OCR confidence for the relevant document. An example of a metadata record is shown in Figure \ref{fig: Metadata record structure}. The overall OCR confidence represents the aggregate confidence score provided by the OCR engine, reflecting the estimated recognition reliability across all pages of the document.

\section{Word Frequency Coverage}
\label{sec:appendix_WordFrequency}

This appendix provides detailed word frequency statistics to complement the analysis mentioned in Section \ref{sec:wordFrequency_and_Coverage}. The Table \ref{tab:word_coverage} summarises the proportion of the corpus accounted for by the most frequent words. Coverage is calculated as the percentage of total word tokens represented by the top 20, 50, and 100 words in Acts and Bills. These figures illustrate the dominance of a small set of high-frequency function words in Sinhala legal texts.

% word coverage table for acts and bills
\begin{table}[h!tb]
\centering
\resizebox{0.45\textwidth}{!}{
\begin{tabular}{lcc} % 4 columns: first left, others stretch
\hline 
\textbf{Number of Word Covered}&
\textbf{Acts} & \textbf{Bills}\\
\hline
Top 20 words & 23.00\% & 23.32\% \\ Top 50 words & 35.04\% & 35.53\% \\ Top 100 words & 45.89\% & 46.39\% \\

\hline
\end{tabular}
}
\caption{Coverage of the most frequent words in Acts and Bills. }
\label{tab:word_coverage}
\end{table}

% Top 10 frequent used words for ACts
\begin{table}[h!tb]
\centering
\resizebox{0.35\textwidth}{!}{
\begin{tabular}{lc} % 4 columns: first left, others stretch
\hline 
\textbf{Most Frequent words in Acts}&
\textbf{Count} \\
\hline
\raisebox{-0.5ex}{
\includegraphics[height=1.5\fontcharht\font`\A]{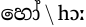}} & 47,559 \\ 
\raisebox{-0.5ex}{
\includegraphics[height=1.5\fontcharht\font`\A]{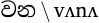}} & 32,380 \\ 
\raisebox{-0.5ex}{
\includegraphics[height=1.5\fontcharht\font`\A]{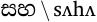}} & 29,775  \\
\raisebox{-0.5ex}{
\includegraphics[height=1.5\fontcharht\font`\A]{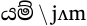}} &  28,440  \\
\raisebox{-0.5ex}{
\includegraphics[height=1.5\fontcharht\font`\A]{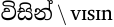}} &  25,494  \\
\raisebox{-0.5ex}{
\includegraphics[height=1.5\fontcharht\font`\A]{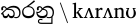}} &  25,105  \\
\raisebox{-0.5ex}{
\includegraphics[height=1.5\fontcharht\font`\A]{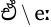}} &   24,947  \\
\raisebox{-0.5ex}{
\includegraphics[height=1.5\fontcharht\font`\A]{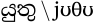}} &   24,027 \\
\raisebox{-0.5ex}{
\includegraphics[height=1.5\fontcharht\font`\A]{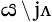}} &    21,079  \\
\raisebox{-0.5ex}{
\includegraphics[height=1.5\fontcharht\font`\A]{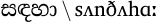}} &   19,933  \\

\hline
\end{tabular}
}
\caption{The top 10 frequent words seen in Acts }
\label{tab:repetitiveWords_Acts}
\end{table}

% Top 10 frequent used words for Bills
\begin{table}[h!tb]
\centering
\resizebox{0.35\textwidth}{!}{
\begin{tabular}{lc} % 4 columns: first left, others stretch
\hline 
\textbf{Most Frequent words in Bills}&
\textbf{Count} \\
\hline
\raisebox{-0.5ex}{
\includegraphics[height=1.5\fontcharht\font`\A]{Sinhala_Words/si_word_ho.pdf}} & 5,489 \\ 
\raisebox{-0.5ex}{
\includegraphics[height=1.5\fontcharht\font`\A]{Sinhala_Words/si_word_saha.pdf}} & 5,309  \\
\raisebox{-0.5ex}{
\includegraphics[height=1.5\fontcharht\font`\A]{Sinhala_Words/si_word_vana.pdf}} &  4,136 \\ 
\raisebox{-0.5ex}{
\includegraphics[height=1.5\fontcharht\font`\A]{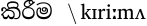}} &  3,776  \\
\raisebox{-0.5ex}{
\includegraphics[height=1.5\fontcharht\font`\A]{Sinhala_Words/si_word_yuthu.pdf}} &  3,622  \\
\raisebox{-0.5ex}{
\includegraphics[height=1.5\fontcharht\font`\A]{Sinhala_Words/si_word_ya.pdf}} &  3,372 \\
\raisebox{-0.5ex}{
\includegraphics[height=1.5\fontcharht\font`\A]{Sinhala_Words/si_word_yam.pdf}} &   3,249 \\
\raisebox{-0.5ex}{
\includegraphics[height=1.5\fontcharht\font`\A]{Sinhala_Words/si_word_visin.pdf}} &   3,235 \\
\raisebox{-0.5ex}{
\includegraphics[height=1.5\fontcharht\font`\A]{Sinhala_Words/si_word_sandaha.pdf}} &     3,124  \\
\raisebox{-0.5ex}{
\includegraphics[height=1.5\fontcharht\font`\A]{Sinhala_Words/si_word_karanu.pdf}} &  2,885  \\

\hline
\end{tabular}
}
\caption{The top 10 frequent words seen in Bills }
\label{tab:repetitiveWords_Bills}
\end{table}

The top 10 frequent words were taken from Acts and Bills separately. They are shown in Table \ref{tab:repetitiveWords_Acts} and Table \ref{tab:repetitiveWords_Bills}. Unlike general Sinhala text, legal text contains complex wordings, but also contains a vast number of repetitive words, which are mostly conjunctions. As shown in the two tables, Acts and Bills mostly contained the same set of frequent words, just in different amounts. Bills mostly contained lower amounts than Acts since the number of Acts in the dataset is higher than that of Bills.

The top 10 most frequently seen bigrams\footnote{A pair of consecutive written units such as letters or words} were also taken. This count was taken as an addition of both Acts and Bills. Words such as \hspace*{-3pt}\raisebox{-0.5ex}{
\includegraphics[height=1.5\fontcharht\font`\A]{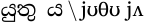}}, \hspace*{-3pt}\raisebox{-0.5ex}{
\includegraphics[height=1.5\fontcharht\font`\A]{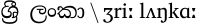}}, \hspace*{-3pt}\raisebox{-0.5ex}{
\includegraphics[height=1.6\fontcharht\font`\A]{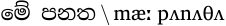}} could be seen to be used frequently across the document. The top 10 bigrams can be seen in Table \ref{tab:bigrams_documents}. 

As mentioned in the Section \ref{sec: calculating_Perplexity_Scores}, these may be the reason legal documents had lower perplexity scores than that of general Sinhala text. These repetitive words and frequent structures are well known to reduce perplexity, as they make it easier for the model to guess the next word.
\cite{yao-etal-2025-understanding}.

 %Top 10 frequently used bigrams for Bills
\begin{table}[h!tb]
\centering
\resizebox{0.35\textwidth}{!}{
\begin{tabular}{lc} % 4 columns: first left, others stretch
\hline 
\textbf{Most Frequent Bigrams}&
\textbf{Count} \\
\hline
\raisebox{-0.5ex}{
\includegraphics[height=1.5\fontcharht\font`\A]{Sinhala_Words/si_word_yuthuya.pdf}} & 17,912 \\ 
\raisebox{-0.5ex}{
\includegraphics[height=1.5\fontcharht\font`\A]{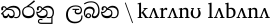}} &  9,004  \\
\raisebox{-0.5ex}{
\includegraphics[height=1.5\fontcharht\font`\A]{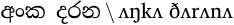}} &  7,814 \\ 
\raisebox{-0.5ex}{
\includegraphics[height=1.5\fontcharht\font`\A]{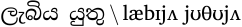}} &  7,453  \\
\raisebox{-0.5ex}{
\includegraphics[height=1.5\fontcharht\font`\A]{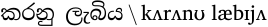}} &     7,268  \\
\raisebox{-0.5ex}{
\includegraphics[height=1.5\fontcharht\font`\A]{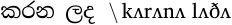}} &  6,860 \\
\raisebox{-0.5ex}{
\includegraphics[height=1.5\fontcharht\font`\A]{Sinhala_Words/si_word_sriLanka.pdf}} &   6,463 \\
\raisebox{-0.5ex}{
\includegraphics[height=1.5\fontcharht\font`\A]{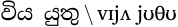}} &   5,566 \\
\raisebox{-0.5ex}{
\includegraphics[height=1.5\fontcharht\font`\A]{Sinhala_Words/si_word_mePanatha.pdf}} &      5,202  \\
\raisebox{-0.5ex}{
\includegraphics[height=1.5\fontcharht\font`\A]{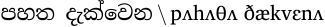}} &  5,089  \\

\hline
\end{tabular}
}
\caption{The top 10 frequent bigrams seen in \textsc{SinhaLegal} }
\label{tab:bigrams_documents}
\end{table}

% Table : Topic modelling results

\begin{table*}[!htb] 
\centering
\resizebox{!}{0.47\textheight}{
\begin{tabular}{|p{4cm}|m{7cm}|c|}
\hline
\textbf{Topic} & \textbf{Word Distributions} & \textbf{Probabilities} \\
\hline

% Topic 1 &
% \raisebox{-0.5ex}{ \includegraphics[height=1.5\fontcharht\font`\A]{latex/Sinhala_Words/si_word_panatha.pdf}}, 
% \raisebox{-0.5ex}{ \includegraphics[height=1.5\fontcharht\font`\A]{latex/Sinhala_Words/si_word_pradhana.pdf}},  
% \raisebox{-0.5ex}{ \includegraphics[height=1.5\fontcharht\font`\A]{latex/Sinhala_Words/si_word_praghapanatha.pdf}}, 
% \raisebox{-0.5ex}{ \includegraphics[height=1.5\fontcharht\font`\A]{latex/Sinhala_Words/si_word_sanshodanaya.pdf}}, 
% \raisebox{-0.5ex}{ \includegraphics[height=1.5\fontcharht\font`\A]{latex/Sinhala_Words/si_word_upawaganthiya.pdf}} 
% & 0.034, 0.018, 0.018,  0.018, 0.017
% \\

\multirow{5}{*}{Topic 1}
& \raisebox{-0.5ex}{\includegraphics[height=1.5\fontcharht\font`\A]{Sinhala_Words/si_word_panatha.pdf}}
& 0.034 \\

& \raisebox{-0.5ex}{\includegraphics[height=1.5\fontcharht\font`\A]{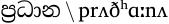}}
& 0.018 \\

& \raisebox{-0.5ex}{\includegraphics[height=1.5\fontcharht\font`\A]{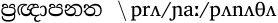}}
& 0.018 \\

& \raisebox{-0.5ex}{\includegraphics[height=1.7\fontcharht\font`\A]{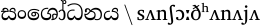}}
& 0.018 \\

& \raisebox{-0.5ex}{\includegraphics[height=1.7\fontcharht\font`\A]{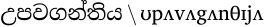}}
& 0.017 \\
\hline

\multirow{5}{*}{Topic 2}
& \raisebox{-0.5ex}{\includegraphics[height=1.7\fontcharht\font`\A]{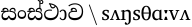}}
& 0.020 \\

& \raisebox{-0.5ex}{\includegraphics[height=1.5\fontcharht\font`\A]{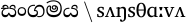}}
& 0.015 \\

& \raisebox{-0.5ex}{\includegraphics[height=1.7\fontcharht\font`\A]{Sinhala_Words/si_word_mahathaa.pdf}}
& 0.012 \\

& \raisebox{-0.5ex}{\includegraphics[height=1.5\fontcharht\font`\A]{Sinhala_Words/si_word_panatha.pdf}}
& 0.012 \\

& \raisebox{-0.5ex}{\includegraphics[height=1.5\fontcharht\font`\A]{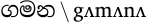}}
& 0.011 \\
\hline

\multirow{5}{*}{Topic 3}
& \raisebox{-0.5ex}{\includegraphics[height=1.7\fontcharht\font`\A]{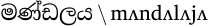}}
& 0.021 \\

& \raisebox{-0.5ex}{\includegraphics[height=1.5\fontcharht\font`\A]{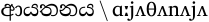}}
& 0.016 \\

& \raisebox{-0.5ex}{\includegraphics[height=1.7\fontcharht\font`\A]{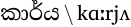}}
& 0.014 \\

& \raisebox{-0.5ex}{\includegraphics[height=1.5\fontcharht\font`\A]{Sinhala_Words/si_word_panatha.pdf}}
& 0.012 \\

& \raisebox{-0.5ex}{\includegraphics[height=1.5\fontcharht\font`\A]{Sinhala_Words/si_word_pradhana.pdf}}
& 0.011 \\
\hline

\multirow{5}{*}{Topic 4}
& \raisebox{-0.5ex}{\includegraphics[height=1.5\fontcharht\font`\A]{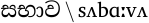}}
& 0.039 \\

& \raisebox{-0.5ex}{\includegraphics[height=1.7\fontcharht\font`\A]{Sinhala_Words/si_word_komishan.pdf}}
& 0.032 \\

& \raisebox{-0.5ex}{\includegraphics[height=1.5\fontcharht\font`\A]{Sinhala_Words/si_word_panatha.pdf}}
& 0.014 \\

& \raisebox{-0.5ex}{\includegraphics[height=1.7\fontcharht\font`\A]{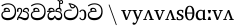}}
& 0.009 \\

& \raisebox{-0.5ex}{\includegraphics[height=1.7\fontcharht\font`\A]{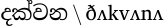}}
& 0.008 \\
\hline

\multirow{5}{*}{Topic 5}
& \raisebox{-0.5ex}{\includegraphics[height=1.7\fontcharht\font`\A]{Sinhala_Words/si_word_sansthawa.pdf}}
& 0.049 \\

& \raisebox{-0.5ex}{\includegraphics[height=1.5\fontcharht\font`\A]{Sinhala_Words/si_word_panatha.pdf}}
& 0.019 \\

& \raisebox{-0.5ex}{\includegraphics[height=1.7\fontcharht\font`\A]{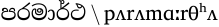}}
& 0.012 \\

& \raisebox{-0.5ex}{\includegraphics[height=1.5\fontcharht\font`\A]{Sinhala_Words/si_word_gamana.pdf}}
& 0.012 \\

& \raisebox{-0.5ex}{\includegraphics[height=1.7\fontcharht\font`\A]{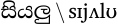}}
& 0.011 \\
\hline

\multirow{5}{*}{Topic 6}
& \raisebox{-0.5ex}{\includegraphics[height=1.7\fontcharht\font`\A]{Sinhala_Words/si_word_chanda.pdf}}
& 0.0019 \\

& \raisebox{-0.5ex}{\includegraphics[height=1.5\fontcharht\font`\A]{Sinhala_Words/si_word_panatha.pdf}}
& 0.010 \\

& \raisebox{-0.5ex}{\includegraphics[height=1.7\fontcharht\font`\A]{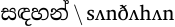}}
& 0.009 \\

& \raisebox{-0.5ex}{\includegraphics[height=1.7\fontcharht\font`\A]{Sinhala_Words/si_word_vishrama.pdf}}
& 0.009 \\

& \raisebox{-0.5ex}{\includegraphics[height=1.7\fontcharht\font`\A]{Sinhala_Words/si_word_upawaganthiya.pdf}}
& 0.008 \\
\hline

\multirow{5}{*}{Topic 7}
& \raisebox{-0.5ex}{\includegraphics[height=1.5\fontcharht\font`\A]{Sinhala_Words/si_word_panatha.pdf}}
& 0.0017 \\

& \raisebox{-0.5ex}{\includegraphics[height=1.7\fontcharht\font`\A]{Sinhala_Words/si_word_adhikaranaya.pdf}}
& 0.013 \\

& \raisebox{-0.5ex}{\includegraphics[height=1.7\fontcharht\font`\A]{Sinhala_Words/si_word_mandalaya.pdf}}
& 0.011 \\

& \raisebox{-0.5ex}{\includegraphics[height=1.7\fontcharht\font`\A]{Sinhala_Words/si_word_karya.pdf}}
& 0.008 \\

& \raisebox{-0.5ex}{\includegraphics[height=1.7\fontcharht\font`\A]{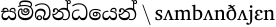}}
& 0.007 \\
\hline

\multirow{5}{*}{Topic 8}
& \raisebox{-0.5ex}{\includegraphics[height=1.5\fontcharht\font`\A]{Sinhala_Words/si_word_panatha.pdf}}
& 0.021 \\

& \raisebox{-0.5ex}{\includegraphics[height=1.5\fontcharht\font`\A]{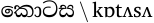}}
& 0.010 \\

& \raisebox{-0.5ex}{\includegraphics[height=1.7\fontcharht\font`\A]{Sinhala_Words/si_word_upawaganthiya.pdf}}
& 0.010 \\

& \raisebox{-0.5ex}{\includegraphics[height=1.7\fontcharht\font`\A]{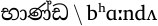}}
& 0.010 \\

& \raisebox{-0.5ex}{\includegraphics[height=1.7\fontcharht\font`\A]{Sinhala_Words/si_word_dakwana.pdf}}
& 0.010 \\
\hline

\multirow{5}{*}{Topic 9}
& \raisebox{-0.5ex}{\includegraphics[height=1.5\fontcharht\font`\A]{Sinhala_Words/si_word_sabhawa.pdf}}
& 0.037 \\

& \raisebox{-0.5ex}{\includegraphics[height=1.7\fontcharht\font`\A]{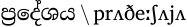}}
& 0.026 \\

& \raisebox{-0.5ex}{\includegraphics[height=1.7\fontcharht\font`\A]{Sinhala_Words/si_word_mandalaya.pdf}}
& 0.015 \\

& \raisebox{-0.5ex}{\includegraphics[height=1.5\fontcharht\font`\A]{Sinhala_Words/si_word_panatha.pdf}}
& 0.012 \\

& \raisebox{-0.5ex}{\includegraphics[height=1.5\fontcharht\font`\A]{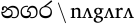}}
& 0.011 \\
\hline

\multirow{5}{*}{Topic 10}
& \raisebox{-0.5ex}{\includegraphics[height=1.5\fontcharht\font`\A]{Sinhala_Words/si_word_panatha.pdf}}
& 0.016 \\

& \raisebox{-0.5ex}{\includegraphics[height=1.7\fontcharht\font`\A]{Sinhala_Words/si_word_mudal.pdf}}
& 0.011 \\

& \raisebox{-0.5ex}{\includegraphics[height=1.7\fontcharht\font`\A]{Sinhala_Words/si_word_upawaganthiya.pdf}}
& 0.009 \\

& \raisebox{-0.5ex}{\includegraphics[height=1.5\fontcharht\font`\A]{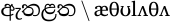}}
& 0.009 \\

& \raisebox{-0.5ex}{\includegraphics[height=1.7\fontcharht\font`\A]{Sinhala_Words/si_word_sambandayen.pdf}}
& 0.008 \\
\hline

\hline
\end{tabular}
}
\caption{The word distributions and their word probabilities for the identified topics from topic modelling.}
\label{tab:topicModelling_wordDistribution}
\end{table*}

\section{Topic Modelling}
\label{sec:appendix_TopicModelling}

Topic modelling was done in order to detect the main themes distributed within the dataset. Before this, the corpus text was tokenised into units. \cite{manning2008introduction}. A list of Sinhala stop words taken from a publicly available repository (mentioned in Section \ref{sec:appendix_TopicModelling}), including conjunctions and other function words, was removed to reduce noise. The list was modified with some common stop words that were also seen in the \textsc{SinhaLegal} dataset.

LDA \cite{blei2003latent} was selected for topic modelling because of its ability to uncover latent thematic structures in large corpora and its interpretability in corpus linguistics. The model was implemented using the Gensim library in Python. After exploratory runs, the number of topics was set to ten, balancing interpretability with coverage.

The revealing topics revealed recurring themes in Sinhala legal discourse. The word distribution for each topic is listed in Table \ref{tab:topicModelling_wordDistribution}. Across multiple topics \raisebox{-0.5ex}{ \includegraphics[height=1.5\fontcharht\font`\A]{Sinhala_Words/si_word_panatha.pdf}} meaning “Act/Law”, emerged as a dominant term, reflecting the centrality of legislative acts in the corpus. 

Other topics highlighted institutional references such as "Council" (\hspace*{-3pt}\raisebox{-0.5ex}{ \includegraphics[height=1.5\fontcharht\font`\A]{Sinhala_Words/si_word_sabhawa.pdf}}), "Court" (\hspace*{-3pt}\raisebox{-0.5ex}{ \includegraphics[height=1.6\fontcharht\font`\A]{Sinhala_Words/si_word_adhikaranaya.pdf}}), "Commission (\hspace*{-3pt}\raisebox{-0.5ex}{ \includegraphics[height=1.6\fontcharht\font`\A]{Sinhala_Words/si_word_komishan.pdf}})" and "Election (\hspace*{-3pt}\raisebox{-0.5ex}{ \includegraphics[height=1.7\fontcharht\font`\A]{Sinhala_Words/si_word_chanda.pdf}})". Themes related to "Money" (\hspace*{-3pt}\raisebox{-0.5ex}{ \includegraphics[height=1.6\fontcharht\font`\A]{Sinhala_Words/si_word_mudal.pdf}}), "Pension" (\hspace*{-3pt}\raisebox{-0.5ex}{ \includegraphics[height=1.6\fontcharht\font`\A]{Sinhala_Words/si_word_vishrama.pdf}}), "Towns" (\hspace*{-3pt}\raisebox{-0.5ex}{ \includegraphics[height=1.5\fontcharht\font`\A]{Sinhala_Words/si_word_nagara.pdf}}) were also to be seen among the listed words. These topics highlight the dominance of legislative references across the corpus, alongside institutional and procedural vocabulary.

\begin{figure*}[!htbp]
  \centering
  \includegraphics[width=2.0\columnwidth]{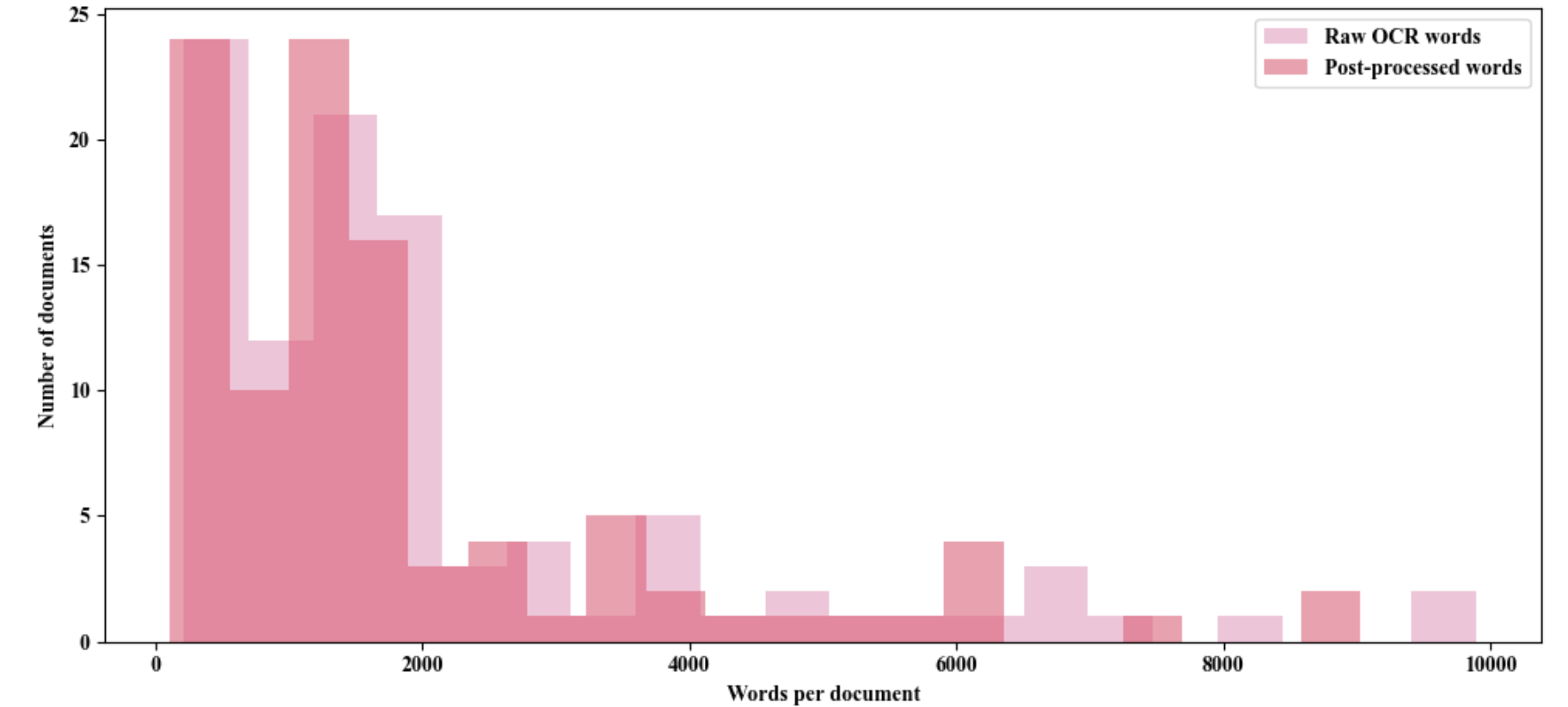}
  \caption{Distribution of words per document before and after OCR}
  \label{fig: distribution of words_ocr}
\end{figure*}

\section{Calculating WER and CER}
\label{sec:appendix_WER_CER}

To calculate the WER and CER, a subset that consisted of 100 legal documents was taken. It contained over 184,000 words and 911,000 characters. Documents vary substantially in length, reflecting the natural diversity of legal texts, with word counts ranging from short to long legal documents. More information regarding the sample subset can be found in Table \ref{tab:sample_statistics}.

\begin{table}[h!tb]
\centering
\resizebox{0.45\textwidth}{!}{
\begin{tabular}{lc} % 3 columns: first left, others stretch
\hline
\textbf{Statistic} & \textbf{Value}  \\
\hline
Documents analysed & 100\\
Total words & 184,011 \\
Average words per document & 1,840\\
Median words per document & 1,295\\
Word count range & 112 - 9,028\\
Total characters & 911,524\\
Average characters per document & 9,115\\
\hline
\end{tabular}
}
\caption{Summary statistics of the evaluated document
subset}
\label{tab:sample_statistics}
\end{table}

Two types of assessments were conducted to provide a fair assessment. This included the original %Character Error Rate%
CER and WER, which include formatting differences and a normalised evaluation focusing solely on content errors. %\raisebox{-0.5ex}{ % Adjust this value if needed for perfect vertical alignment

First, an original evaluation was conducted where the raw OCR texts were compared directly against the post-processed texts. The randomly selected 100 documents were matched with the same files to the raw OCR files to calculate the WER and CER. The results showed a CER value of 24.07\% and a WER value of 26.87\%. 

To assess the content-level fidelity independently of formatting, the post-processed texts were normalised \cite{lucas_icdar_2005}. The normalisation included collapsing multiple consecutive spaces into a single space, reducing multiple consecutive line breaks to a single line break, removing leading and trailing white spaces from each line, and discarding empty lines. After the normalisation, the CER and WER were recalculated using the same document subset. As shown in Table \ref{tab:WER_and_CER_results}, normalisation reduced WER while leaving CER largely unchanged.

%WER and CER table
\begin{table}[h!tb]
\centering
\resizebox{0.45\textwidth}{!}{
\begin{tabular}{lcc} % 3 columns: first left, others stretch
\hline
\textbf{Metric} & \textbf{Original Evaluation} & \textbf{Normalized Evaluation} \\
\hline
CER & 24.07\% & 24.06\% \\
WER & 26.87\% & 23.44\% \\
\hline
\end{tabular}
}
\caption{Comparison of OCR evaluation metrics before and after text normalisation.}
\label{tab:WER_and_CER_results}
\end{table}

The minor change in CER suggests that the removal of formatting doesn't affect the result of the calculation and that the earlier results of the CER were mostly accurate, while the WER dropped by nearly 3\%. 

\section{Structural Analysis}
\label{sec:appendix_structuralAnalysis}
 
Other than calculating the WER and CER, a structural comparison was done. This was illustrated using a single document from the 100 documents already sampled to provide a concrete example (Document name: acts\_1988-04 21\_Evidence\_Amendment\_si.txt).  It was seen that the post-processing effectively reduced redundant line breaks and spaces, which improves consistency and readability for future tasks. 

\begin{table}[h!tb]
\centering
\resizebox{0.45\textwidth}{!}{
\begin{tabular}{lccc} % 4 columns: first left, rest auto-stretch
\hline
\textbf{Feature} & \textbf{Raw OCR} & \textbf{Post-processed} & \textbf{Difference} \\
\hline
Line breaks & 39 & 22 & 17 \\
Spaces & 173 & 94 & 79 \\
\hline
\end{tabular}
}
\caption{Structural comparison of raw OCR and post-processed text, showing the number of line breaks and spaces removed during post-processing.}

\label{tab:structural_comparison}
\end{table}

To qualitatively illustrate OCR errors and post-processing corrections, a character-level comparison of the same representative document was done. The Table \ref{tab:char_difference} highlights the removal of spurious digits, punctuation, redundant line breaks, and OCR-induced noise in Sinhala characters. 

% Character level difference between raw and ORC text: table 
\begin{table}[h!tb]
\centering
\resizebox{0.45\textwidth}{!}{
\begin{tabular}{lc} 
\hline
\textbf{Operation} & \textbf{Characters Added or Removed}  \\
\hline
Removed & 2\\
Removed & 8 \\
Removed & 1\\
Removed & \raisebox{-0.5ex}{
\includegraphics[height=1.5\fontcharht\font`\A]{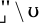}} \\
Removed & \raisebox{-0.5ex}{
\includegraphics[height=1.1\fontcharht\font`\A]{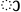}} \\
Removed & \raisebox{-0.5ex}{
\includegraphics[height=1.4\fontcharht\font`\A]{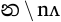}} \\
Added & \textbackslash n\\
Added & \raisebox{-0.5ex}{
\includegraphics[height=1.5\fontcharht\font`\A]{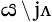}}\\
\hline
\end{tabular}
}
\caption{The character-level differences between a raw OCR and post-processed text in a document}
\label{tab:char_difference}
\end{table}

% \begin{figure*}[!htbp]
%   \centering
%   \includegraphics[width=2.0\columnwidth]{latex/Diagrams/WordsPerDocument.PNG}
%   \caption{Distribution of words per document before and after OCR}
%   \label{fig: distribution of words_ocr}
% \end{figure*}

In addition to the document-level structural illustration, a corpus-level structural analysis was conducted over the full subset of 100 documents to quantify the overall impact of post-processing on document length characteristics.
At the document level, the average number of words per document decreased from 2,113 in the raw OCR to 1,840 after post-processing. Similarly, the average character count decreased from 12,506 to 11,007 characters per document. Removing unnecessary characters, stamp content, and footers results in a decrease in the number of words.  Table \ref{tab:structural_comparison_words_characters} includes more information regarding the change of words and characters during the process. Figure \ref{fig: distribution of words_ocr} also demonstrates the distribution of the words per document before and after OCR.

\begin{table}[h!tb]
\centering
\resizebox{0.45\textwidth}{!}{
\begin{tabular}{lcc} % 3 columns: first left, others stretch
\hline
\textbf{Statistic} & \textbf{Raw OCR} & \textbf{Post-processed} \\
\hline
Total words & 211,317& 184,011\\
Total characters & 1,250,623 & 1,100,695 \\
Average words per document & 2,113 & 1,840\\
Average characters per document & 1,2506 & 11007 \\
Word count range & 213--9891 & 112--9028 \\
\hline
\end{tabular}
}
\caption{Values comparing raw OCR and post-processed documents on the word and character count.}
\label{tab:structural_comparison_words_characters}
\end{table}

Taken together, these results demonstrate that the post-processing approach improves structural consistency throughout the corpus. While Table \ref{tab:structural_comparison} and the qualitative examples highlight localised corrections within a single document, the aggregate statistics confirm that similar structural improvements are achieved across the entire dataset.

\begin{table*}[t]
\centering

% Word level corrections: table %
\resizebox{0.95\textwidth}{!}{
\begin{tabular}{lcc}
\hline
\textbf{Scanned Document View} & \textbf{OCR Extracted Text} & \textbf{Corrected Text} \\ \hline

\includegraphics[width=5cm]{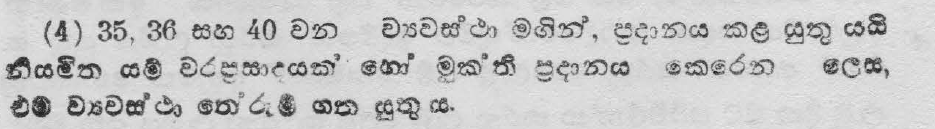} &
\includegraphics[width=5cm]{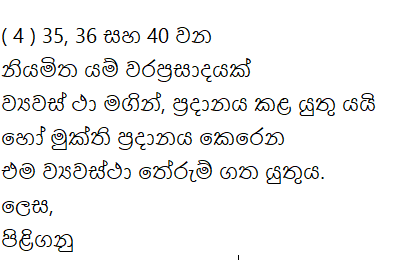} &
\includegraphics[width=5cm]{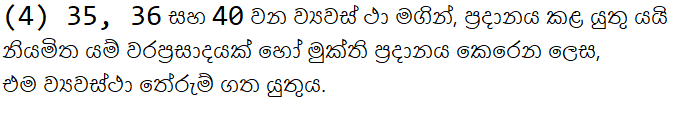} \\ \hline

\includegraphics[width=5cm]{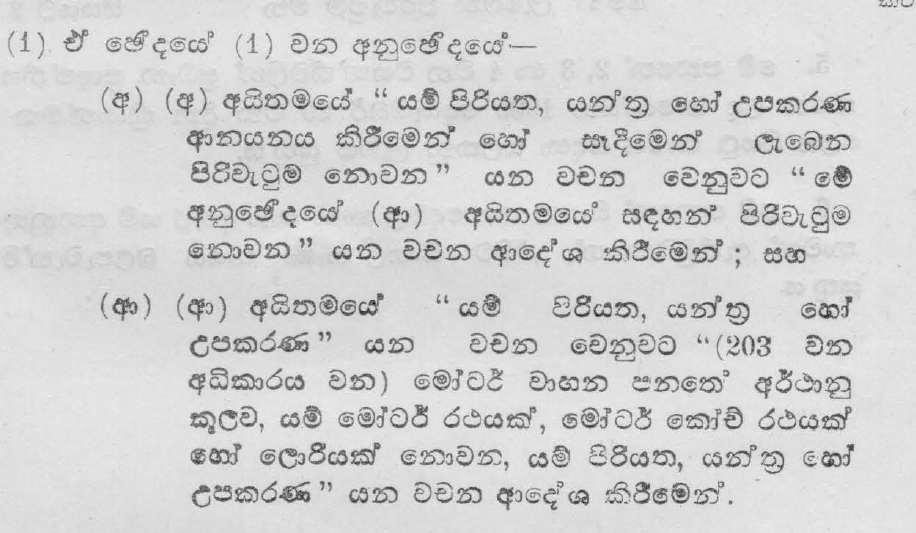} &
\includegraphics[width=5cm]{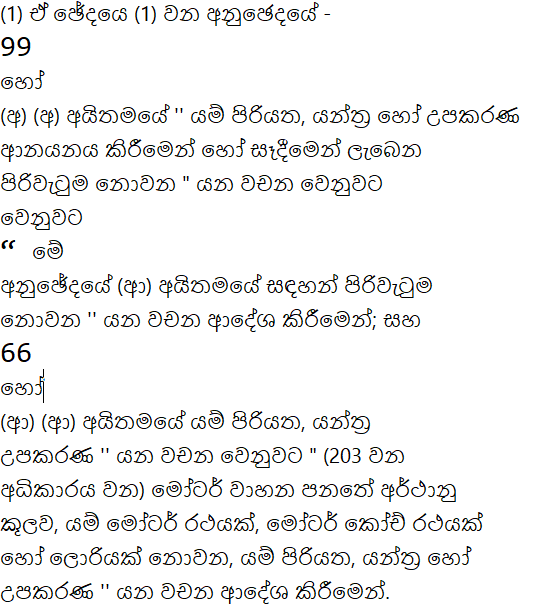} &
\includegraphics[width=5cm]{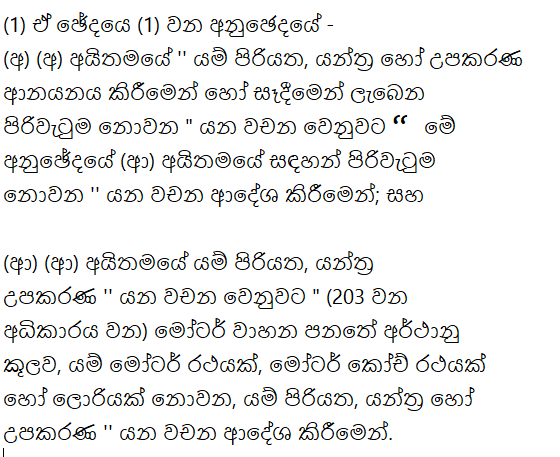} \\ \hline

\includegraphics[width=5cm]{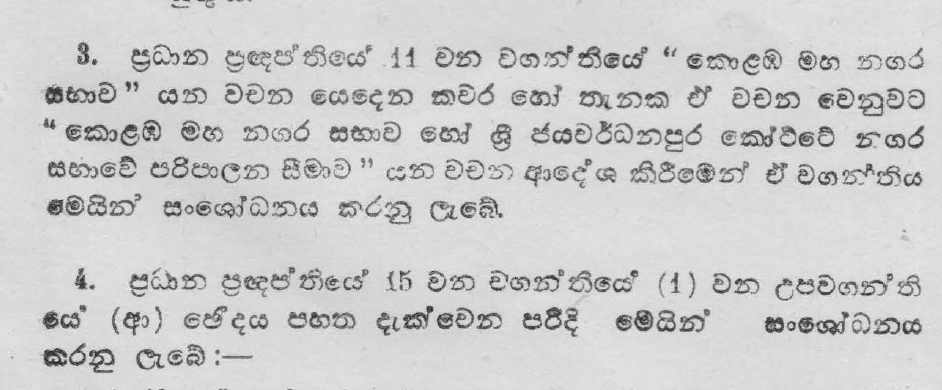} &
\includegraphics[width=5cm]{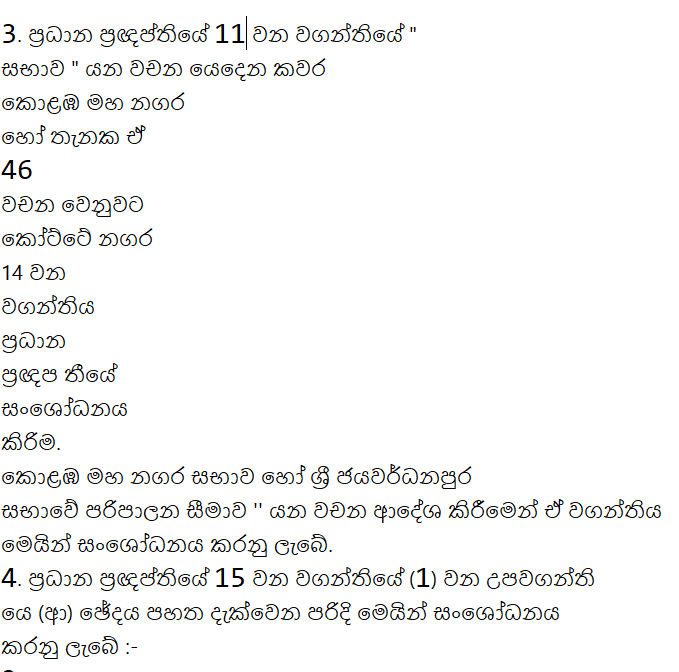} &
\includegraphics[width=5cm]{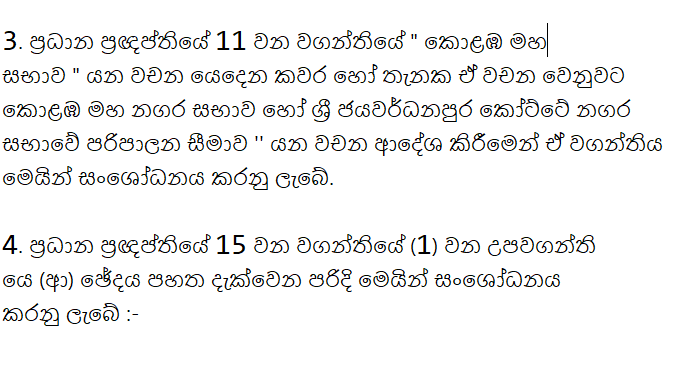} \\ \hline
\end{tabular}}

\caption{Example of scanned PDF, OCR text, and corrected text for word-level corrections.}
\label{tab:word_level_corrections}

\vspace{1cm} 

% Spacing corrections: table %
\resizebox{0.95\textwidth}{!}{
\begin{tabular}{lcc}
\hline
\textbf{Scanned Document View} & \textbf{OCR Extracted Text} & \textbf{Corrected Text} \\ \hline

\includegraphics[width=5cm]{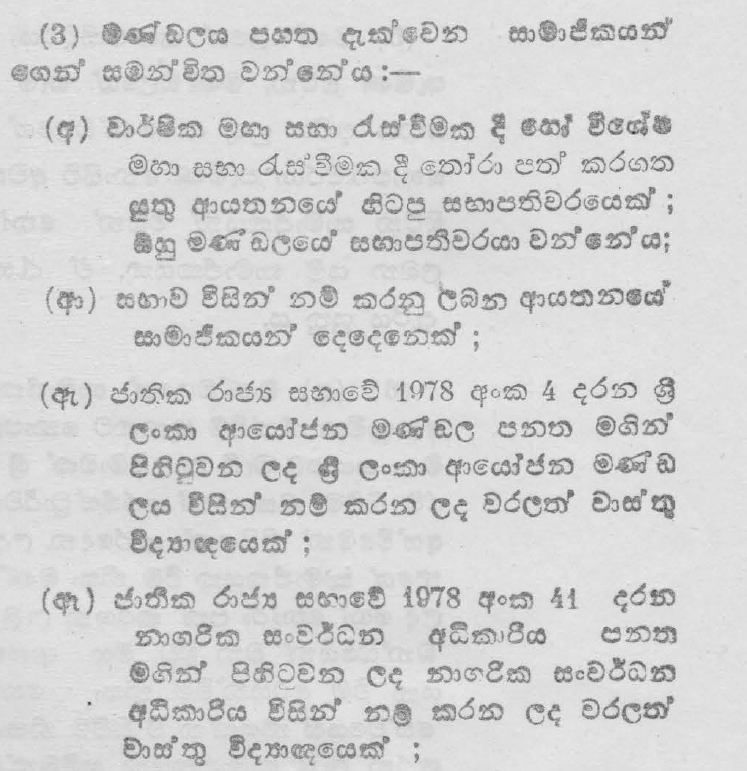} &
\includegraphics[width=5cm]{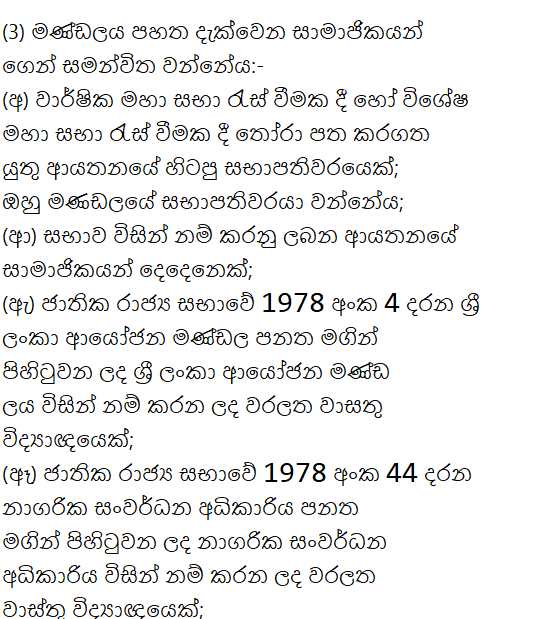} &
\includegraphics[width=5cm]{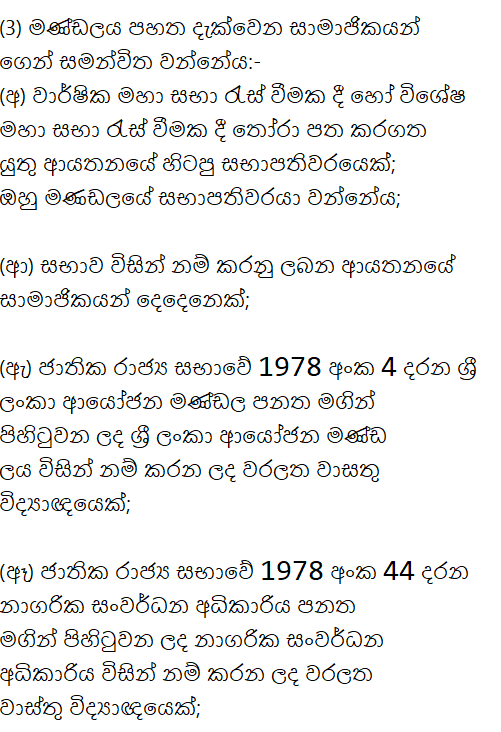}\\ \hline

\end{tabular}}
\caption{Example of scanned PDF, extracted text, and corrected text for spacing errors.}
\label{tab:spacing_corrections}

\end{table*}

\end{document}